\documentclass{article}

\usepackage[preprint]{neurips_2024}

\usepackage[utf8]{inputenc} 
\usepackage[T1]{fontenc}    
\usepackage{hyperref}       
\usepackage{url}            
\usepackage{booktabs}       
\usepackage{amsfonts}       
\usepackage{nicefrac}       
\usepackage{microtype}      
\usepackage{xcolor}         

\usepackage{amsmath,amsfonts,amssymb,amsthm}
\usepackage{bbm}
\usepackage{graphicx}
\usepackage{multirow}
\usepackage{wrapfig}
\usepackage{tikz}
\usepackage{algorithm}
\usepackage{algpseudocode}

\newcommand{\R}{\mathbb{R}}

\newcommand{\I}{\mathbbm{1}} 
\newcommand{\norm}[1]{\left\lVert#1\right\rVert}

\newcommand{\E}{\mathbb{E}}

\newcommand{\ones}{\mathbf{1}}

\newcommand{\col}{\mathop{\text{Col}}}

\newcommand{\ie}{i.e.}
\renewcommand{\hat}{\widehat}
\renewcommand{\tilde}{\widetilde}

\newtheorem{thm}{Theorem}

\newtheorem{lem}[thm]{Lemma} 
\newtheorem{cor}[thm]{Corollary}

\newtheorem{rem}{Remark}

\newtheorem{defn}{Definition}
\newtheorem{ass}{Assumption}

\makeatletter
  \g@addto@macro \normalsize{%
    \setlength\abovedisplayskip{4pt plus 0pt minus 0pt}%
    \setlength\belowdisplayskip{4pt plus 0pt minus 0pt}}%
\makeatother

\title{On the Minimax Regret in Online Ranking \\ with Top-k Feedback}

\author{%
  Mingyuan Zhang\thanks{This work is based on a dissertation submitted by MZ under AT's supervision in partial fulfillment of the requirements for the degree of Bachelor of Science (Honors Statistics) at the University of Michigan.} \\
  University of Pennsylvania\\
  Philadelphia, PA 19104 \\
  \texttt{myz@seas.upenn.edu} \\
  \And
  Ambuj Tewari \\
  University of Michigan, Ann Arbor\\
  Ann Arbor, MI 48109 \\
  \texttt{tewaria@umich.edu} \\
}

\begin{document}

\maketitle

\begin{abstract}
  In online ranking, a learning algorithm sequentially ranks a set of items and receives feedback on its ranking in the form of relevance scores. Since obtaining relevance scores typically involves human annotation, it is of great interest to consider a partial feedback setting where feedback is restricted to the top-$k$ items in the rankings.
  \citet{Chaudhuri17} developed a framework to analyze online ranking algorithms with top-$k$ feedback. A key element in their work was the use of techniques from partial monitoring. In this work, we further investigate online ranking with top-$k$ feedback and solve some open problems posed by \citet{Chaudhuri17}. We provide a full characterization of minimax regret rates with the top-$k$ feedback model for all $k$ and for the following ranking performance measures: Pairwise Loss, Discounted Cumulative Gain, and Precision@n. In addition, we give an efficient algorithm that achieves the minimax regret rate for Precision@n.
\end{abstract}

\section{Introduction}

Ranking problems arise in many applications including search engines, recommendation systems, and online advertising (see, e.g., the book by \citet{Liu11}). The output space in ranking problems consists of permutations of objects. Given the true relevance scores of the objects, the accuracy of a ranked list is judged using ranking measures, such as Pairwise Loss (PL), Discounted Cumulative Gain (DCG), and Precision@n (P@n). Many ranking algorithms are {\em offline}, i.e., they are designed to operate on the entire data in a single batch. However, interest in {\em online} algorithms, i.e., those that process the data incrementally, is rising for a number of reasons. First, online algorithms often require less computation and storage. Second, many applications, especially on the web, produce ongoing streams of data making them excellent candidates for applying online algorithms. Third, basic online algorithms, such as the ones developed in this work, make excellent starting points for developing more sophisticated online algorithms that can deal with {\em non-stationarity}. Non-stationarity is a major issue in ranking problems since user preferences can easily change over time.

The basic full feedback setting assumes that the relevance scores, typically obtained via human annotation, provide the correct feedback for each item in the ranking. Since the output in ranking problems is a permutation over a potentially large set of objects, it becomes practically impossible to get full feedback from human annotators. Therefore, researchers have looked into {\em weak supervision} or {\em partial feedback} settings where the correct relevance score is only partially revealed to the learning algorithm. For example, \citet{Chaudhuri17} developed a model for online ranking with a particular case of partial feedback called {\em top-$k$ feedback}. In this model, the online ranking problem is cast as an online partial monitoring game between a learner and an oblivious adversary (who generates a sequence of outcomes before the game begins), played over $T$ rounds.\footnote{Some other problems that can be cast as partial monitoring games are multi-armed bandits \citep{Auer03}, and dynamic pricing \citep{Kleinberg03}.} At each round, the learner outputs a ranking of objects whose quality with respect to the true relevance scores of the objects, is judged by some ranking measure. However, the learner receives limited feedback at the end of each round: only the relevance scores of the top-$k$ ranked objects are revealed to the learner. Here, $k \le m$ (in practice $k \ll m$) and $m$ is the number of objects. 

The goal of the learner is to minimize its regret. The goal of regret analysis is to compute the upper bounds of the regret of explicit algorithms. If lower bounds on regret that match the upper bounds up to constants can be derived, then the {\em minimax regret} is identified, again up to constants. Previous work considered two settings: \emph{non-contextual} (objects to be ranked are fixed) and \emph{contextual} (objects to be ranked vary and get encoded as a context, typically in the form of a feature vector). Our focus in this work will be on the non-contextual setting where six ranking measures have been studied: PL, DCG, P@n, and their normalized versions Area Under Curve (AUC), Normalized Discounted Cumulative Gain (NDCG) and Average Precision (AP). \citet{Chaudhuri17} showed that the minimax regret rates with the top-$k$ feedback model for PL, DCG and P@n are upper bounded by $O(T^{2/3})$ for all $1 \le k \le m$. In particular, for $k=1$, the minimax regret rates for PL and DCG are $\Theta(T^{2/3})$. Moreover, for $k=1$, the minimax regret rates for AUC, NDCG, and AP are $\Theta(T)$. One of the open questions, as described by \citet{Chaudhuri17}, is to find the minimax regret rates for $k>1$ for the six ranking measures.

It is worth noting that the top-$k$ feedback model is neither full feedback (where the adversary's move is uniquely determined by the feedback) nor bandit feedback (where the loss is determined by the feedback); the model falls under the framework of \textit{partial monitoring} \citep{Cesa06}. Recent advances in classification of finite partial-monitoring games have shown that the minimax regret of any such game is $0$, $\Theta(T^{1/2})$, $\Theta(T^{2/3})$, or $\Omega(T)$, and is governed by two important properties: \emph{global observability} and \emph{local observability} \citep{Bartok14}. In particular, \citet{Bartok14} gave an almost complete classification of all finite partial-monitoring games by identifying four regimes: trivial, easy, hard, and hopeless games, which correspond to the four minimax regret rates mentioned before, respectively. What was left from the classification is the set of games in oblivious adversarial settings with degenerate actions which are never optimal themselves, but can provide useful information. \citet{Lattimore18,LattimoreS19} finished the characterization of the minimax regret for all partial monitoring games.

\textbf{Our contributions:} We establish the minimax regret rates for all values of $k$, i.e., $1 \le k \le m$ and for ranking measures PL, DCG, and P@n. We obtain these results by showing that the properties of global observability and local observability hold in the appropriate cases. In addition, we provide an algorithm based on the \textsc{NeighborhoodWatch2} algorithm of \citet{Lattimore18}. Our algorithm achieves the minimax rate for P@n and has per-round time complexity polynomial in $m$ (for any fixed $n$).

\section{Notations and Problem Setup}
We defer all proofs to the appendix.

Our problem formulation follows that of \citet{Chaudhuri17}. Let $\{e_i\}$ denote the standard basis.
Let $\ones$ denote the vector of all ones. The fixed $m$ objects to be ranked are $[m] := \{1,...,m\}$. The permutation $\sigma$ maps from ranks to objects, and its inverse $\sigma^{-1}$ maps from objects to ranks. Specifically, $\sigma(i) = j$ means that object $j$ is ranked $i$ and $\sigma^{-1}(i) = j$ means that object $i$ is ranked $j$. The relevance vector $R \in \{0,1\}^m$ represents relevance for each object. $R(i)$, $i$-th component of $R$, is the relevance for object $i$. Sometimes, relevance values can be multi-graded, i.e., can take values other than $0$ or $1$. However, in this work, we only study binary relevance.

The learner can choose from $m!$ actions $\{ \sigma \mid \sigma : [m] \to [m] \text{ is bijective} \}$ while the adversary can choose from $2^m$ outcomes $\{0,1\}^m$. We use subscript $t$ exclusively to denote time $t$, so $\sigma_t$ is the action the learner chooses at round $t$ and $R_t$ is the outcome the adversary chooses at round $t$.

In a game $G$, the learner and the adversary play over $T$ rounds. We consider an {\em oblivious} adversary who chooses all the relevance vectors $R_t$ ahead of the game (but they are not revealed to the learner at that point).\footnote{Note that a stochastic adversary who draws i.i.d. relevance vectors $R_t$ is a special case of an oblivious adversary.} In each round $t$, the learner predicts a permutation (ranking) $\sigma_t$ according to a (possibly randomized) strategy $\pi$. The performance of $\sigma_t$ is judged against $R_t$ by some ranking (loss) measure $RL$. At the end of round $t$, only the relevance scores of the top-$k$ ranked objects ($R_t (\sigma_t (1)),...,R_t (\sigma_t (k))$) are revealed to the learner. Therefore, the learner knows neither $R_t$ (as in the full information game) nor $RL(\sigma_t, R_t)$ (as in the bandit game). The goal of the learner is to minimize the expected regret (where the expectation is over any randomness in the learner's moves $\sigma_t$) defined as the difference in the realized loss and the loss of the best-fixed action in hindsight:
\begin{align}
\mathcal{R}_T(\pi, R_1, ...R_T) := \E_{\sigma_1, ..., \sigma_T} \Big[\sum_{t=1}^{T} RL(\sigma_t, R_t) \Big] - \min_\sigma \sum_{t=1}^{T} RL(\sigma, R_t) \, .
\end{align}
When the ranking measure is a gain, we can negate the gain function so that it becomes a loss function. The worst-case regret of a learner's strategy is its maximum regret over all choices of $R_1,...,R_T$. The minimax regret is the minimum worst-case regret over all strategies of the learner:
\begin{align}
\mathcal{R}_{T}^{*} (G) = \inf_{\pi} \max_{R_1,...,R_T} \mathcal{R}_T(\pi, R_1, ...R_T) \, ,
\end{align}
where $\pi$ is the learner's strategy to generate $\sigma_1,...,\sigma_T$.

\section{Ranking Measures} \label{sec:ranking_measures}

We are interested in ranking measures that can be expressed in the form of $f(\sigma) \cdot R$ where $f : \R^m \to \R^m$, is composed of $m$ copies of a univariate monotonically non-decreasing scalar-valued function $f^s : \R \to \R$. We say that $f^s$ is monotonically non-decreasing if and only if $\sigma^{-1}(i) > \sigma^{-1}(j)$ implies $f^s(\sigma^{-1}(i)) \ge f^s(\sigma^{-1}(j))$. The monotonic non-increasing is defined analogously. Then, $f(\sigma)$ can be written as
\begin{align*}
f(\sigma) = [f^s (\sigma^{-1}(1)), ..., f^s (\sigma^{-1}(m))] \, .
\end{align*}

The definitions of ranking measures that we are going to study in this work are the following.

\textbf{Pairwise Loss (PL) and Sum Loss (SL)}
\begin{align}
PL(\sigma, R) = \sum_{i=1}^{m} \sum_{j=1}^{m} \I(\sigma^{-1} (i) < \sigma^{-1} (j)) \I(R (i) < R (j)) \, .
\end{align}
\begin{align}
SL(\sigma, R) = \sum_{i=1}^{m} \sigma^{-1} (i) R(i) \, .
\end{align}
It has been shown that the regret under PL is equal to the regret under SL (Section 2 of \cite{Ailon14}). Therefore, we can study SL instead of PL. The minimax regret rate for PL is the same as that for SL:
\begin{align} \label{eq:pl_sl}
\sum_{t=1}^{T} PL(\sigma_t, R_t) - \sum_{t=1}^{T} PL(\sigma, R_t) = \sum_{t=1}^{T} SL(\sigma_t, R_t) - \sum_{t=1}^{T} SL(\sigma, R_t) \, .
\end{align}
Although we cannot express PL in the form of $f(\sigma) \cdot R$, we can do that for SL with $f(\sigma) = \sigma^{-1} = [\sigma^{-1}(1), ..., \sigma^{-1}(m)]$.

\textbf{Discounted Cumulative Gain (DCG)}
\begin{align}
DCG(\sigma, R) = \sum_{i=1}^{m} \frac{R(i)}{\log_2 (1+\sigma^{-1} (i))} \, .
\end{align}
Negated DCG can be expressed in the form of $f(\sigma) \cdot R$ with
\begin{align*}
f(\sigma) = [-\frac{1}{\log_2 (1+\sigma^{-1} (1))}, ..., -\frac{1}{\log_2 (1+\sigma^{-1} (m))}] \, .
\end{align*}

\textbf{Precision@n Gain (P@n)}
\begin{align}
\label{eqn:p@n}
P@n(\sigma, R) = \sum_{i=1}^{m} \I(\sigma^{-1} (i) \le n) R(i) \, .
\end{align}
Negated P@n can also be expressed in the form of $f(\sigma) \cdot R$ with
\begin{align*}
f(\sigma) = [-\I(\sigma^{-1} (1) \le n), ..., -\I(\sigma^{-1} (m) \le n)] \, .
\end{align*}

\begin{rem}
There are reasons why we are interested in this linear (in $R$) form of ranking measures. The algorithms that establish upper bounds for the minimax regret rates require the construction of an unbiased estimator of the difference vector between two loss vectors that two different actions incur \citep{Bartok14, Lattimore18}. The nonlinear (in $R$) form of ranking measures (including AUC, NDCG, and AP) makes such a construction extremely hard (see details in \citet{Chaudhuri17}).
\end{rem}

\section{Summary of Results}
We first summarize our main results before delving into technical details. We remind the reader that we assume that the adversary is oblivious and the time horizon is $T$. We operate under a non-contextual top-$k$ feedback model with $m$ objects.\footnote{`Non-contextual' means that there is no side information/features/context associated with the $m$ objects being ranked.}
\begin{enumerate}
    \item 
    The minimax regret rate for PL, SL and DCG is $\Theta(T^{2/3})$ for $k=1,2,...,m-2$ and is $\Theta(T^{1/2})$ for $k=m-1,m$.
    \item
    The minimax regret rate for P@n is $\Theta(T^{1/2})$ for $1 \le k \le m$.
    \item
    The minimax regret rate for P@n can be achieved by an efficient algorithm that needs only $O(\mathrm{poly}(m))$ rounds and has per-round time complexity $O(\mathrm{poly}(m))$.
\end{enumerate}

\section{Finite Partial Monitoring Games} \label{sec:finite_partial_monitoring}
Our results on minimax regret rates are developed based on the theory for general finite partial monitoring games developed by \citet{Bartok14,Lattimore18,LattimoreS19}. Before presenting our results, it is necessary to reproduce the relevant definitions and notations as in \citet{Bartok14},  \citet{Chaudhuri17}, and \citet{Lattimore18}. For the sake of easy understanding, we adapt the definitions and notation to our setting.

\subsection{A Quick Review of Finite Partial Monitoring Games}

Recall that in the top-$k$ feedback model, there are $m!$ actions and $2^m$ outcomes (because we only consider binary relevance). Without loss of generality, we fix an ordering $(\sigma_i)_{1\le i \le m!}$ of all the actions and an ordering $(R_j)_{1\le j \le 2^m}$ of all the outcomes. Note that the subscripts in $\sigma_i$ and $R_j$ refer to the place in these fixed ordering and do not refer to time points in the game as in $\sigma_t$. It will be clear from the context whether we are referring to a place in the ordering or to a time point in the game. A game with ranking measure $RL$ and top $k \ (1 \le k \le m)$ feedback can be defined by a pair of \textit{loss matrix} and \textit{feedback matrix}. The \textit{loss matrix} is denoted by $L \in \R^{m! \times 2^m}$ with rows corresponding to actions and columns corresponding to outcomes. $L_{i,j}$ is the loss the learner suffers when the learner chooses action $\sigma_i$, and the adversary chooses outcome $R_j$, i.e., $L_{i,j} = RL(\sigma_i, R_j)$. The \textit{feedback matrix} is denoted by $H$ of size $m! \times 2^m$ with rows corresponding to actions and columns corresponding to outcomes. $H_{i,j}$ is the feedback the learner gets when the learner chooses action $\sigma_i$, and the adversary chooses outcome $R_j$, i.e., $H_{i,j} = (R_j (\sigma_i (1)),...,R_j (\sigma_i (k)))$.

Loss matrix $L$ and feedback matrix $H$ together determine the difficulty of a game. In the following, we will introduce some definitions to help understand the underlying structures of $L$ and $H$.

Let $l_i$ denote the column vector consisting of the $i$-th row of $L$. It is also called the loss vector for action $i$. Let $\Delta$ be the probability simplex in $\R^{2^m}$, that is, $\Delta = \{ p \in \R^{2^m} : p \ge 0, \ones^\top p = 1 \}$ where the inequality between vector is to be interpreted component-wise. Elements of $\Delta$ can be treated as \textit{opponent strategies} as they are distributions of overall outcomes. With loss vectors and $\Delta$, we can then define what it means for a learner's action to be optimal.
\begin{defn} [Optimal action]
Learner's action $\sigma_i$ is said to be \textbf{optimal} under $p \in \Delta$ if $l_i \cdot p \le l_j \cdot p$ for all $1 \le j \le m!$. That is, $\sigma_i$ has an expected loss not greater than that of any other learner's actions under $p$.
\end{defn}
Identifying opponent strategies under which an action is optimal gives the \textit{cell decomposition} of $\Delta$.
\begin{defn} [Cell decomposition]
For learner's action $\sigma_i$, $1 \le i \le m!$, its \textbf{cell} is defined to be $C_i = \{ p \in \Delta: l_i \cdot p \le l_j \cdot p, \forall 1 \le j \le m! \}$. Then $\{C_1, ..., C_{m!}\}$ forms the \textbf{cell decomposition} of $\Delta$.
\end{defn}
It is easy to see that each cell is either empty or is a closed polytope. Based on the properties of different cells, we can classify corresponding actions as follows.
\begin{defn} [Classification of actions]
Action $\sigma_i$ is called \textbf{dominated} if $C_i = \emptyset$. Action $\sigma_i$ is called \textbf{nondominated} if $C_i \neq \emptyset$. Action $\sigma_i$ is called \textbf{degenerate} if it is nondominated and there exists action $\sigma_j$ such that $C_i \subsetneq C_j$. Action $\sigma_i$ is called \textbf{Pareto-optimal} if it is nondominated and not degenerate.
\end{defn}
Dominated actions are never optimal. Cells of Pareto-optimal actions have $(2^m - 1)$ dimensions, while those of degenerate actions have dimensions strictly less than $(2^m - 1)$.

Sometimes two actions might have the same loss vector, and we will call them duplicate actions. Formally, action $\sigma_i$ is called \textbf{duplicate} if there exists action $\sigma_j \neq \sigma_i$ such that $l_i = l_j$. If actions $\sigma_i$ and $\sigma_j$ are duplicates of each other, one might think of removing one of them without loss of generality. Unfortunately, this will not work. Even though $\sigma_i$ and $\sigma_j$ have the same loss vector, they might have different feedback. Thus removing one of them might lead to a loss of information that the learner could have received.

Next, we introduce the concept of \textit{neighbors} defined in terms of Pareto-optimal actions.
\begin{defn} [Neighbors] \label{defn:neighbors}
Two Pareto-optimal actions $\sigma_i$ and $\sigma_j$ are \textbf{neighboring actions} if $C_i \cap C_j$ has dimension $(2^m - 2)$. The \textbf{neighborhood action set} of two neighboring actions $\sigma_i$ and $\sigma_j$ is defined as $N_{i,j}^{+} = \{k' : 1 \le k' \le m!, C_i \cap C_j \subseteq C_{k'} \}$.
\end{defn}
All of the definitions above are with respect to the loss matrix $L$. The structure of $L$ (\ie, the number of each type of actions) certainly plays an important role in determining the difficulty of a game. (For example, if a game has only one Pareto-optimal action, then simply playing the Pareto-optimal action in each round leads to zero regret.) However, that is only half of the story. In the other half, we will see the feedback matrix $H$ determines how easily we can identify optimal actions.

In the following, we will turn our attention to the feedback matrix $H$. Recall that $H_{i,j}$ is the feedback the learner gets when the learner plays action $\sigma_i$, and the adversary plays outcome $R_j$. Consider the $i$-th row of $H$, which is all possible feedback the learner could receive when playing action $i$. We want to infer what outcome the adversary chose from the feedback. Thus, the feedback itself does not matter; what matters is the number of distinct symbols in the $i$-th row of $H$. This will determine how easily we can differentiate among outcomes. Therefore, we will use \textit{signal matrices} to standardize the feedback matrix $H$.
\begin{defn} [Signal matrix]
\label{defn:signal-matrix}
Recall that in top-$k$ feedback model, the feedback matrix has $2^k$ distinct symbols $\{0,1\}^{k}$. Fix an enumeration $s_1, ..., s_{2^k}$ of these symbols. Then the \textbf{signal matrix} $S_i \in \{0,1\}^{2^k \times 2^m}$, corresponding to action $\sigma_i$, is defined as $(S_i)_{l,l'} = \I(H_{i,l'} = s_l)$.
\end{defn}
At this point, one might attempt to construct unbiased estimators for loss vectors for all actions and then apply algorithms like Exp3 \citep{Auer1998}. Unfortunately, this approach will not work in this setting. There are easy counterexamples (see Exhibit 1 in Appendix H of \citet{Lattimore18}). Another approach is to construct unbiased estimators for differences between loss vectors. The idea is that we do not need to estimate the loss itself; instead, it suffices to estimate how an action performs with respect to the optimal action in order to control the regret. It turns out this idea indeed works. The following two definitions capture the difficulty with which we can construct an unbiased estimator for loss vector difference.
\begin{defn} [Global observability] \label{defn:go}
A pair of actions $\sigma_i$ and $\sigma_j$ is called \textbf{globally observable} if $l_i - l_j \in \oplus_{1 \le k' \le m!} \col(S_{k'}^\top)$, where $\col$ refers to column space. The \textbf{global observability} condition holds if every pair of neighboring actions is globally observable.
\end{defn}
\begin{defn} [Local observability] \label{defn:lo}
A pair of neighboring actions $\sigma_i$ and $\sigma_j$ is called \textbf{locally observable} if $l_i - l_j \in \oplus_{k' \in N_{i,j}^{+}} \col(S_{k'}^\top)$. The \textbf{local observability} condition holds if every pair of neighboring actions is locally observable.
\end{defn}
Global observability means that the loss vector difference can be estimated using feedback from all actions, while local observability means that it can be estimated using just feedback from the neighborhood action set. Clearly, local observability is a stronger condition, and it implies global observability.

We note that the above two definitions are given in \citet{Bartok14}. Later, when \citet{Lattimore18} extended \citet{Bartok14}'s work, they proposed different (but equivalent) definitions of \textit{global observability} and \textit{local observability}. We reproduce as follows.
\begin{defn} [Alternative definitions of global observability and local observability] \label{defn:alter_go_lo}
Let $\Sigma = \{ \sigma \mid \sigma : [m] \to [m] \text{is bijective}\}$. Let $\mathcal{H}$ denote the set of symbols in $H$. A pair of actions $\sigma_i$ and $\sigma_j$ is called \textbf{globally observable} if there exists a function $f : \Sigma \times \mathcal{H} \to \R$ such that
\[
\sum_{k'=1}^{m!} f(\sigma_{k'}, H_{k',l'}) = L_{i,l'} - L_{j, l'} \quad \text{for all } 1 \le l' \le 2^m \, .
\]
A pair of actions $\sigma_i$ and $\sigma_j$ are \textbf{locally observable} if in addition to the above they are neighbors and $f(\sigma_{k'}, \cdot) = 0$ when $k' \notin N_{i,j}^{+}$. Again, the \textbf{global observability} condition holds if every pair of neighboring actions is globally observable, and the \textbf{local observability} condition holds if every pair of neighboring actions is locally observable.
\end{defn}
\begin{lem} \label{lem:alter_go_lo}
The alternative definitions of global observability and local observability (Definition \ref{defn:alter_go_lo}) are equivalent to the original definitions of global observability and local observability (Definition \ref{defn:go} and Definition \ref{defn:lo}), respectively.
\end{lem}
To explain intuitively, note that $\sigma_{k'}$ and $H_{k', l'}$ contain the same information as $S_{k'}$ and $e_{l'}$ because observing $H_{k', l'}$ is equivalent to observing $S_{k'} e_{l'}$. The latter can be seen as a one-hot coding vector for the feedback. Since the two sets of definitions are equivalent, we choose to use the one that is more convenient in the context.

\subsection{Classification Theorem for Finite Partial Monitoring Games}
To make this work self-contained, we will state the important result that we use from the theory of finite partial monitoring games. The following theorem provides a full classification of all finite partial monitoring games into four categories.
\begin{thm} \label{thm:classification_finite_partial_monitoring} [Theorem 2 in \citet{Bartok14}, Theorem 2 in \citet{Lattimore18}, and Theorem 9 in \citet{LattimoreS19}]
The minimax regret rate of partial monitoring game $G = (L, H)$ satisfies
\begin{align*}
R_T^*(G) = \begin{cases}
0, & \text{if $G$ has no pairs of neighboring actions}; \\
\Theta(T^{1/2}), & \text{if $G$ is locally observable and has neighboring actions}; \\
\Theta(T^{2/3}), & \text{if $G$ is globally observable, but not locally observable}; \\
\Omega(T), & \text{otherwise}.
\end{cases}
\end{align*}
\end{thm}
This theorem involves upper and lower bounds for each of the four categories. Several papers \citep{Piccolboni2001, Antos2013, Cesa06, Bartok14, Lattimore18,LattimoreS19} contribute to this theorem. In particular, \citet{Bartok14} summarizes and gives a nearly complete classification theorem. However, they failed to deal with degenerate games (\ie, the game that has degenerate or duplicate actions). This is important for us since, as we shall see later, the game for P@n contains duplicate actions. Fortunately, \citet{Lattimore18} filled this gap in the literature. Later, \citet{LattimoreS19} eliminated the logarithmic dependence on $T$ for locally observable games that appeared in earlier work.

Intuitively, a game is `trivial' (zero regret) if no learning is needed. This only happens when there is one Pareto-optimal action or all Pareto-optimal actions are duplicates. A game is `easy' ($T^{1/2}$ regret) if a learner needs to play an action with a small price in order to gain information (for loss vector difference estimation). A game is `hard' ($T^{2/3}$ regret) if a learner needs to play an action with a heavy price in order to gain information. A game is `hopeless' (linear regret) if no information can be obtained to determine optimal actions (for example, two actions with the same feedback).

With this classification theorem, it suffices for us to show the local or global observability conditions in order to establish minimax regret rates.

\section{Minimax Regret Rates for PL, SL and DCG} \label{sec:pl_sl_dcg}
We first show the minimax regret rate for a family of ranking loss measures that satisfy the following assumption.
\begin{ass}[Strict increasing property] \label{ass:rl_strictly_increasing}
The ranking loss measure $RL(\sigma, R)$ can be expressed in the form $f(\sigma) \cdot R$ where $f : \R^m \to \R^m$, is composed of $m$ copies of a univariate strictly increasing scalar-valued function $f^s : \R \to \R$, that is, $\sigma^{-1}(i) > \sigma^{-1}(j)$ implies $f^s(\sigma^{-1}(i)) > f^s(\sigma^{-1}(j))$.
\end{ass}
As mentioned in Section \ref{sec:ranking_measures}, SL satisfies Assumption \ref{ass:rl_strictly_increasing} with $f(\sigma) = \sigma^{-1} = [\sigma^{-1}(1), ..., \sigma^{-1}(m)]$. The negated DCG, which is a ranking loss measure, also satisfies Assumption \ref{ass:rl_strictly_increasing} with $f(\sigma) = [-\frac{1}{\log_2 (1+\sigma^{-1} (1))}, ..., -\frac{1}{\log_2 (1+\sigma^{-1} (m))}]$.

In the following, assume the ranking loss measure $RL$ satisfies Assumption \ref{ass:rl_strictly_increasing} unless otherwise stated. We first identify the classification of actions and their corresponding cell decomposition.

\begin{lem}[Classification of actions for $RL$ satisfying Assumption \ref{ass:rl_strictly_increasing}] \label{lem:rl_pareto_optimal_actions}
For $RL$ that satisfies Assumption \ref{ass:rl_strictly_increasing}, each of the learner's actions $\sigma_i$ is Pareto-optimal.
\end{lem}
Determining whether the game is locally observable requires knowing all neighboring action pairs. We now characterize neighboring action pairs for $RL$.
\begin{lem}[Neighboring action pair for $RL$ satisfying Assumption \ref{ass:rl_strictly_increasing}] \label{lem:rl_neighboring_action_pairs}
A pair of actions $\sigma_i$ and $\sigma_j$ is a neighboring action pair if and only if there is exactly one pair of objects $\{a,b\}$, whose positions differ in $\sigma_i$ and $\sigma_j$, such that $a$ is placed just before $b$ in $\sigma_i$, and $b$ is placed just before $a$ in $\sigma_j$.
\end{lem}
\begin{rem} \label{rem:rl_loss_diff_observation}
From Lemma \ref{lem:rl_neighboring_action_pairs}, neighboring action pair $\{\sigma_i, \sigma_j\}$ has the form: $\sigma_i(k') = a, \sigma_i(k'+1) = b, \sigma_j(k') = b, \sigma_j(k'+1) = a$ for some $k'$, and $\sigma_i(l) = \sigma_j(l), \forall l \neq k', k'+1$, for objects $a$ and $b$. By properties of $RL$, we can calculate the $R_t$ entry of $l_i - l_j$ (the entry that corresponds to $R_t$) as $RL(\sigma_i, R_t) - RL(\sigma_j, R_t) = R_t \cdot (f(\sigma_i) - f(\sigma_j)) = R_t(a) \cdot (f^s(\sigma_{i}^{-1}(a)) - f^s(\sigma_{j}^{-1}(a))) + R_t(b) \cdot (f^s(\sigma_{i}^{-1}(b)) - f^s(\sigma_{j}^{-1}(b))) = R_t(a) \cdot (f^s(k') - f^s(k'+1)) + R_t(b) \cdot (f^s(k'+1) - f^s(k'))$. We can see that $l_i - l_j$ contains $2^{m-1}$ nonzero entries, of which $2^{m-2}$ entries are $f^s(k') - f^s(k'+1)$ and $2^{m-2}$ entries are $f^s(k'+1) - f^s(k')$. Moreover, if $R_t(a) = 1$ and $R_t(b) = 0$ for the $R_t$ relevance, then the $R_t$ entry of $l_i - l_j$ is $f^s(k') - f^s(k'+1)$. If $R_t(a) = 0$ and $R_t(b) = 1$ for the $R_t$ relevance, then the $R_t$ entry of $l_i - l_j$ is $f^s(k'+1) - f^s(k')$. If $R_t(a) = R_t(b)$ for the $R_t$ relevance, then the $R_t$ entry of $l_i - l_j$ is $0$.
\end{rem}
Once we know what a neighboring action pair is, we can characterize the corresponding neighborhood action set.
\begin{lem}[Neighborhood action set for $RL$ satisfying Assumption \ref{ass:rl_strictly_increasing}] \label{lem:rl_neighborhood_action_set}
For neighboring action pair $\{\sigma_i, \sigma_j\}$, the neighborhood action set is $N_{i,j}^{+} = \{i,j\}$, so $\oplus_{k \in N_{i,j}^{+}} \col(S_{k}^\top) = \col(S_{i}^\top) \oplus \col(S_{j}^\top)$.
\end{lem}
We are now ready to state the first important theorem in this work.
\begin{thm}[Local observability for $RL$ satisfying Assumption \ref{ass:rl_strictly_increasing}] \label{thm:rl}
Under top-$k$ feedback model with $m$ objects, with respect to loss matrix $L$ and feedback matrix $H$ for $RL$ satisfying Assumption \ref{ass:rl_strictly_increasing}, the local observability fails for $k=1,...,m-2$ and holds for $k=m-1, m$.
\end{thm}
Since SL satisfies Assumption \ref{ass:rl_strictly_increasing}, we have the following corollary from Theorem \ref{thm:rl}.
\begin{cor}[Local observability for SL] \label{cor:pl}
With respect to loss matrix $L$ and feedback matrix $H$ for SL, the local observability fails for $k=1,...,m-2$ and holds for $k=m-1, m$.
\end{cor}
\textbf{Minimax regret rates for SL and PL.} Theorem 1 in section 2.4 and the discussion in section 2.5 of \citet{Chaudhuri17} have shown that for SL, the global observability holds for all $1 \le k \le m$. Combining our Theorem \ref{thm:rl} and chaining with Theorem \ref{thm:classification_finite_partial_monitoring}, we immediately have the minimax regret for SL:
\begin{align*}
\mathcal{R}_{T}^{*} =
\begin{cases}
\Theta(T^{2/3}), \quad k = 1,...,m-2 \\
\Theta(T^{1/2}), \quad k = m-1, m \\
\end{cases} \, .
\end{align*}
By Equation (\ref{eq:pl_sl}), PL has exactly the same minimax regret rates as SL.

\textbf{Discussion.}
Corollary \ref{cor:pl} shows that this game is hard for almost all values of $k$. In particular, since, in reality, small values of $k$ are more interesting, it rules out the possibility of better regret for practically interesting cases for $k$. We also note that \citet{Chaudhuri17} showed the failure of local observability only for $k=1$.

As for the time complexity, \citet{Chaudhuri17} provided an efficient (polynomial of $m$ time) algorithm for PL and SL for values of $k$ when global observability holds, so we have an efficient algorithm for $k=1,2,...,m-2$. For $k=m$, \citet{Suehiro12} and \citet{Ailon14} have already shown efficient algorithms. The only case left out is $k=m-1$. Such a large value of $k$ is not interesting in practice, so we do not pursue this question.

Similarly, since negated DCG also satisfies Assumption \ref{ass:rl_strictly_increasing}, we have the following corollary from Theorem \ref{thm:rl}.
\begin{cor}[Local observability for DCG] \label{cor:dcg}
With respect to loss matrix $L$ and feedback matrix $H$ for DCG, the local observability fails for $k=1,...,m-2$ and holds for $k=m-1, m$.
\end{cor}
\textbf{Minimax regret rate for DCG.} Corollary 10 of \citet{Chaudhuri17} showed that for DCG, the minimax regret rate is $O(T^{2/3})$ for $1 \le k \le m$. Combining with our Corollary \ref{cor:dcg} and chaining with Theorem \ref{thm:classification_finite_partial_monitoring}, we immediately have the minimax regret for DCG:
\begin{align*}
\mathcal{R}_{T}^{*} =
\begin{cases}
\Theta(T^{2/3}), \quad k = 1,...,m-2 \\
\Theta(T^{1/2}), \quad k = m-1, m \\
\end{cases} \, .
\end{align*}

\textbf{Discussion.}
Corollary \ref{cor:dcg} generalizes the results in \citet{Chaudhuri17} that showed local observability fails only for $k=1$, and rules out the possibility of better regret for values of $k$ that are practically interesting. Also, there are efficient algorithms for $k=1,2,...,m-2$ \citep{Chaudhuri17} and for $k=m$ \citep{Suehiro12,Ailon14}. Again, we are not interested in designing an efficient algorithm for $k=m-1$.

\section{Minimax Regret Rate for P@n} \label{sec:p@n}
The negated P@n does not satisfy Assumption \ref{ass:rl_strictly_increasing} because $f^s$ is not strictly increasing (see Eq. \eqref{eqn:p@n}), so Theorem \ref{thm:rl} does not apply to negated P@n.

In the following, the ranking loss measure is negated P@n unless otherwise stated. To establish a minimax regret rate for P@n, we first need to identify the classification of actions and their corresponding cell decomposition.
\begin{lem}[Classification of actions for P@n] \label{lem:p@n_pareto_optimal_actions}
For negated P@n, each of the learner's actions $\sigma_i$ is Pareto-optimal. 
\end{lem}
Next, we characterize neighboring action pairs for negated P@n.
\begin{lem}[Neighboring action pairs for P@n] \label{lem:p@n_neighboring_action_pairs}
For negated P@n, a pair of learner's actions $\{\sigma_i, \sigma_j\}$ is a neighboring action pair if and only if there is exactly one pair of objects $\{a,b\}$ such that $a \in A_i$, $a \in B_j$, $b \in B_i$, and $b \in A_j$, where $A_i = \{ a : \sigma_{i}^{-1}(a) \le n\}$, $B_i = \{ b : \sigma_{i}^{-1}(b) > n \}$, and $A_j$ and $B_j$ are defined similarly.
\end{lem}
\begin{rem} \label{rem:p@n_loss_diff_observation}
From Lemma \ref{lem:p@n_neighboring_action_pairs}, for neighboring action pair $\{\sigma_i, \sigma_j\}$, we know there is exactly one pair of objects $\{a,b\}$ such that $a \in A_i$, $a \in B_j$, $b \in B_i$, and $b \in A_j$, where $A_i, A_j, B_i, B_j$ are defined as in Lemma \ref{lem:p@n_neighboring_action_pairs}. Using the definition of negated P@n, we can see that $l_i - l_j$ contains $2^{m-1}$ nonzero entries, of which $2^{m-2}$ entries are $1$ and $2^{m-2}$ entries are $-1$. Moreover, if $R_s(a) = 1$ and $R_s(b) = 0$ for the $s$-th ($1 \leq s \leq 2^m$) relevance, then the $s$-th entry of $l_i - l_j$ is $-1$. If $R_s(a) = 0$ and $R_s(b) = 1$ for the $s$-th ($1 \leq s \leq 2^m$) relevance, then the $s$-th entry of $l_i - l_j$ is $1$. If $R_s(a) = R_s(b)$ for the $s$-th ($1 \leq s \leq 2^m$) relevance, then the $s$-th entry of $l_i - l_j$ is $0$.
\end{rem}
Then, we characterize the neighborhood action set for a neighboring action pair.
\begin{lem}[Neiborhood action set for P@n] \label{lem:p@n_neighborhood_action_set}
For neighboring action pair $\{\sigma_i, \sigma_j\}$, the neighborhood action set is $N_{i,j}^{+} = \{k: 1 \le k \le m!, l_k = l_i \text{ or } l_k = l_j\}$.
\end{lem}
\begin{rem} \label{rem:p@n_duplicate_actions}
Negated P@n says that it only matters the way of partitioning $m$ objects into 2 sets $A$ and $B$ as in Lemma \ref{lem:p@n_pareto_optimal_actions}. For a fixed partition $A$ and $B$, we can permute objects within $A$ and within $B$, and all such permutations give the same loss vector and the same cell. Thus, there are duplicate actions in P@n, but no degenerate actions.
\end{rem}
We are prepared to state the local observability theorem for P@n. The proof uses the same technique as that in the proof of Theorem \ref{thm:rl}.
\begin{thm}[Local observability for P@n] \label{thm:p@n}
For fixed $n$ such that $1 \le n \le m$, with respect to loss matrix $L$ and feedback matrix $H$ for P@n, the local observability holds for all $1 \le k \le m$.
\end{thm}

\textbf{Minimax regret rate for P@n.} Note that this game contains many duplicate actions (but no degenerate actions) since P@n only cares about objects ranked in the top $n$ position, irrespective of the order. The minimax regret does not directly follow from Theorem 2 of \citet{Bartok14}. However, \citet{Lattimore18,LattimoreS19} have proved that locally observable games enjoy $\Theta(T^{1/2})$ minimax regret, regardless of the existence of duplicate actions. This shows the minimax regret for P@n is
\[
{R}_{T}^{*} = \Theta(T^{1/2}) \quad \text{for } 1 \le k \le m \, .
\]

\textbf{Discussion.} We note that \citet{Chaudhuri17} only showed $O(T^{2/3})$ regret rates for P@n, so this result gives improvements over all values of $k$, including the practically relevant cases when $k$ is small. In the next section, we will also give an efficient algorithm that realizes this regret rate.

\section{Efficient Algorithm for Obtaining Minimax Regret Rate for P@n} \label{sec:p@n_algo}
\citet{Lattimore18} showed an algorithm \textsc{NeighborhoodWatch2} that achieves  $\tilde {\Theta}(T^{1/2})$ minimax regret for all finite partial monitoring games with local observability, including games with duplicate or degenerate actions.\footnote{$\tilde {\Theta}(\cdot)$ indicates growth up to logarithmic factors.}
In particular, they showed that \textsc{NeighborhoodWatch2} achieves $\Theta(T^{1/2})$ minimax regret when there are no degenerate actions.\footnote{This result appeared in the arXiv version of \citet{Lattimore18}. See \url{https://arxiv.org/abs/1805.09247} for \citet{Lattimore18arXiv}.}
However, directly applying this algorithm to P@n would be intractable, since the algorithm has to spend $\Omega(\mathrm{poly}(K))$ time per round, where the number of actions $K$ equals $m!$ in our setting with P@n.

We provide a modification before applying the algorithm \textsc{NeighborhoodWatch2} so that it spends only $O(\mathrm{poly}(m))$ time per round and obtains a minimax regret rate of $\Theta(T^{1/2})$. Thus, it is more efficient.
We note that since top-$k$ (for $k > 1$) feedback contains strictly more information than top-1 feedback does, it suffices to give an efficient algorithm for P@n with top-1 feedback which we will show in the following.
We first give a high-level idea of why we can significantly reduce the time complexity from exponential in $m$ to polynomial in $m$. It has to do with the structure of the game for P@n.

Lemma \ref{lem:p@n_neighboring_action_pairs} says that P@n only cares about how action $\sigma$ partitions $[m]$ into sets $A$ and $B$; the order of objects within $A$ (or $B$) does not matter. Furthermore, each ordering of objects in $A$ and $B$ corresponds to a unique action. Therefore, based on loss vectors, we can define equivalent classes over $m!$ actions such that all actions within a class share the same loss vector. In other words, each class collects actions duplicated to each other. A simple calculation shows all classes have the same number of actions, $n! (m-n)!$, and there are $\binom{m}{n}$ classes. Note that $\binom{m}{n}$ is $O(m^n)$ for any fixed $n$, a polynomial of $m$. In real applications, $n$ is usually very small, such as $1, 3, 5$.
In each of the equivalent classes, all the actions have the same partition of $[m]$ into sets $A$ and $B$, where all objects in $A$ are ranked before objects in $B$. For top-1 feedback setting, the algorithm only receives the relevance for the object ranked at the top. Therefore, in a class, the algorithm only needs to determine which object from $A$ to be placed at the top position. Clearly, there are just $n$ choices as there are $n$ objects in $A$, so we reduce the number of actions to consider in each class from $n! (m-n)!$ to $n$. Note that this reduction does not incur any loss of information. This is the key idea to simplify the time complexity. We only need to keep track of a distribution to sample from $n \binom{m}{n}$ (a polynomial of $m$ for any fixed $n$) actions, instead of sampling from $m!$ actions.

Let $\mathcal{C}$ be the set of those $n \binom{m}{n}$ actions defined above. Let $\mathcal{A}$ be an arbitrary largest subset of Pareto-optimal actions from $\mathcal{C}$ such that $\mathcal{A}$ does not contain actions that are duplicates of each other. Note that $\mid\mathcal{A}\mid = \binom{m}{n}$ and $\mathcal{A}$ contain an action from each of the equivalent classes. Let $\mathcal{D} = \mathcal{C} \setminus \mathcal{A}$. For action $a$, let $N_a$ be the set of actions consisting of $a$ and $a$'s neighbors.
To make this section self-contained, we include the algorithm \textsc{NeighborhoodWatch2} in \citet{Lattimore18} with some changes so that it is consistent with our notations. See Algorithm \ref{alg:nw2}.\footnote{The original \textsc{NeighborhoodWatch2} algorithm, which this algorithm is based on, requires a \textsc{Redistribute} function at line 6 to handle degenerate actions. Since there are no degenerate actions in this game, we leave it out.}

By the alternative definition of local observability, there exists a function $v^{ab} : \Sigma \times \mathcal{H} \to \R$ for each pair of neighboring actions $a,b$ such that the requirement in Definition \ref{defn:alter_go_lo} is satisfied. For notational convenience, let $v^{aa} = 0$ for all action $a$. Define $V = \max_{a,b} \lVert v^{ab} \rVert_\infty$. Since both $\Sigma$ and $\mathcal{H}$ are finite sets, $\lVert v^{ab} \rVert_\infty$ is just $\max_{\sigma \in \Sigma, s \in \mathcal{H}} \mid v^{ab} (\sigma, s)\mid$. The following lemma shows $\lVert v^{ab} \rVert_\infty \le 4$ for suitable choice of $v^{ab}$, so $V$ can be upper bounded by $4$.
\begin{lem}[Upper bound for $\lVert v^{ab} \rVert_\infty$] \label{lem:nw2_V}
For each pair of neighboring actions $a,b$ for the ranking loss measure (negated) P@n, there exists a function $v^{ab} : \Sigma \times \mathcal{H} \to \R$ such that Definition \ref{defn:alter_go_lo} is satisfied and moreover, $\lVert v^{ab} \rVert_\infty = \max_{\sigma \in \Sigma, s \in \mathcal{H}} \mid v^{ab} (\sigma, s)\mid$ can be upper bounded by $4$.
\end{lem}

\begin{algorithm}[bt]
\begin{algorithmic}[1]
\State \textbf{Input\,\,} $L$, $H$, $\eta$, $\gamma$
\For{$t = 1,\ldots,T$}
\State For $a,k \in \mathcal{C}$, let\\
~~~~~~~~~~$\displaystyle Q_{tka} = \I_{\mathcal{A}}(k) \frac{\I_{N_k\cap \mathcal{A}}(a)\exp\left(-\eta \sum_{s=1}^{t-1} \tilde Z_{ska} \right)}{\sum_{b \in N_k\cap \mathcal{A}} \exp\left(-\eta \sum_{s=1}^{t-1} \tilde Z_{skb}\right)} + \I_{\mathcal{D}}(k) \frac{\I_{\mathcal{A}}(a)}{\mid\mathcal{A}\mid}$
\State Find distribution $\tilde P_t$ such that $\tilde P_t^\top = \tilde P_t^\top Q_t$
\State Compute $P_t = (1 - \gamma) \tilde P_t + \frac{\gamma}{\mid\mathcal{C}\mid} \ones$
\State Sample $A_t \sim P_t$ and receive feedback $\Phi_t$
\State Compute loss-difference estimators for each $k \in \mathcal{A}$ and $a \in N_k \cap \mathcal{A}$:\\
~~~~~~~~~~$\hat Z_{tka} = \frac{\tilde P_{tk} v^{ak}(A_t, \Phi_t)}{P_{tA_t}} \, ,$\\
~~~~~~~~~~$\beta_{tka} = \eta V^2 \sum_{b \in N_{ak}^+} \frac{\tilde P_{tk}^2}{P_{tb}} \, ,$ and\\
~~~~~~~~~~$\tilde Z_{tka} = \hat Z_{tka} - \beta_{tka}$
\EndFor
\end{algorithmic}
\caption{\textsc{NeighborhoodWatch2}}\label{alg:nw2}
\end{algorithm}
Below, we provide an upper bound on the regret when running Algorithm \ref{alg:nw2} for the top-$k$ feedback model with P@n.

\begin{thm} [\citet{Lattimore18arXiv,Lattimore18}] \label{thm:p@n_algo}
Let $K = \mid\mathcal{C}\mid = n \binom{m}{n}$. For top-1 feedback model with P@n, suppose Algorithm \ref{alg:nw2} is run on $G = (L,H)$ with $\eta = \frac{1}{V}\sqrt{\log(K)/T}$ and $\gamma = \eta K V$. Then 
\[
\E[\mathcal{R}_{T}] \le O\Big(\frac{K V}{\epsilon_{G}} \sqrt{T \log (K)}\Big) \, ,
\]
where $\epsilon_G$ is a constant specific to the game $G$, not depending on $T$, and $\frac{V}{\epsilon_G} \le 8m$. Using a loose bound $\binom{m}{n} \le m^n$, we have
\[
\E[\mathcal{R}_{T}] \le O\Big(n m^{n+1} \sqrt{T (\log(n) + n\log(m)) }\Big) \, .
\]
So the regret is upper bounded by $O(\mathrm{poly}(m) \cdot T^{1/2})$ for any fixed $n$.
Moreover, the time complexity in each round is $O(\mathrm{poly}(m))$.
Since top-$k$ (for $k > 1$) feedback contains strictly more information than top-1 feedback does, this result applies to the general top-$k$ feedback model with P@n as well.
\end{thm}

Before proving this theorem, the following lemma shows that $\frac{1}{\epsilon_G}$ can be defined as $2m$ in this setting.
\begin{lem} [Lemma 6 in \citet{Bartok14}, and Lemma 5 in \citet{Lattimore18}] \label{lem:local_regret} \label{lem:nw2_eps_G}
There exists a constant $\epsilon_G > 0$, depending only on $G$, such that for all $c,d \in \mathcal{A}$ and $u \in C_d$ there exists $e \in N_c \cap \mathcal{A}$ with
\[
(l_c - l_d) \cdot u \le \frac{1}{\epsilon_G} (l_c - l_e) \cdot u \, .
\]
Moreover, $\frac{1}{\epsilon_G}$ can be taken as $2m$.
\end{lem}

\textbf{Proof of Theorem \ref{thm:p@n_algo}.}
\begin{proof}
It is easy to see from Algorithm \ref{alg:nw2} that the time complexity in each round is $O(\mathrm{poly}(K)) = O(\mathrm{poly}(m))$. Lemma \ref{lem:nw2_eps_G} shows $\frac{1}{\epsilon_G} = 2m$. From Lemma \ref{lem:nw2_V}, we have $V = \max_{a,b} \lVert v^{ab} \rVert_\infty \le 4$. Then $\frac{V}{\epsilon_G} \le 8m$. The remaining proof follows \citet{Lattimore18arXiv,Lattimore18}.
\end{proof}

\textbf{Discussion.}
We have provided an upper bound on the regret in terms of $m,n,T$. Yet, doing the same for a lower bound is tricky (a known lower bound is $\Omega(T^{1/2})$), so we leave it as future work. \citet{LattimoreS20} has some discussion on lower bounds for locally observable partial monitoring games in terms of the number of actions and the number of outcomes. We refer interested readers to check for further details.

The regret bound for \textsc{NeighborhoodWatch2} involves a game-dependent constant $\frac{1}{\epsilon_G}$ which can be arbitrarily large \citep{Lattimore18}, even though in our setting, we can upper bound it by $2m$.
In the follow-up work, \citet{LattimoreS19} established new regret bounds that are independent of arbitrarily large game-dependent constants. However, their approach is non-constructive.
Then, \citet{LattimoreS20} provided an algorithm so that for locally observable games without degenerate actions (as is the case in our current setting), its regret upper bound matches the best-known information-theoretical upper bound shown in \citet{LattimoreS19}.
When running their algorithm with our current setting, the regret upper bound is $O(K^{3/2}\sqrt{T \log(K)})$, where $K = n \binom{m}{n}$. Comparing it with our bound $O(mK\sqrt{T \log(K)})$ shown in Theorem \ref{thm:p@n_algo}, we can see that our bound is better in our current setting.

Recently, new algorithms based on mirror-decent and information-directed sampling have been proposed for partial monitoring games \citep{TsuchiyaHS20,Lattimore021a,Lattimore22}. They could potentially be applied in our current setting as well.

\section{Conclusion}

In this work, we have successfully closed one of the most interesting open questions proposed by \citet{Chaudhuri17}: we have established a full characterization of minimax regret rates with top-$k$ feedback model for all $k$ for ranking measures Pairwise Loss (PL), Discounted Cumulative Gain (DCG) and Precision@n Gain (P@n).

For PL and DCG, we have improved the results in \citet{Chaudhuri17} and ruled out the possibility of better regret for values of $k$ that are practically interesting. For P@n, which is widely used in learning to rank community, we have shown a surprisingly good regret of $\Theta(T^{1/2})$ for all $k$, which improved the original regret $O(T^{2/3})$ in \citet{Chaudhuri17}. Moreover, we have provided an efficient algorithm that achieves this regret rate.




\newpage


\bibliographystyle{plainnat}

\begin{thebibliography}{18}
\providecommand{\natexlab}[1]{#1}
\providecommand{\url}[1]{\texttt{#1}}
\expandafter\ifx\csname urlstyle\endcsname\relax
  \providecommand{\doi}[1]{doi: #1}\else
  \providecommand{\doi}{doi: \begingroup \urlstyle{rm}\Url}\fi

\bibitem[Ailon(2014)]{Ailon14}
Nir Ailon.
\newblock Improved bounds for online learning over the permutahedron and other ranking polytopes.
\newblock In \emph{Proceedings of the Seventeenth International Conference on Artificial Intelligence and Statistics, {AISTATS} 2014}, volume~33 of \emph{{JMLR} Workshop and Conference Proceedings}, pages 29--37. JMLR.org, 2014.

\bibitem[Antos et~al.(2013)Antos, Bart{\'{o}}k, P{\'{a}}l, and Szepesv{\'{a}}ri]{Antos2013}
Andr{\'{a}}s Antos, G{\'{a}}bor Bart{\'{o}}k, D{\'{a}}vid P{\'{a}}l, and Csaba Szepesv{\'{a}}ri.
\newblock Toward a classification of finite partial-monitoring games.
\newblock \emph{Theor. Comput. Sci.}, 473:\penalty0 77--99, 2013.

\bibitem[Auer et~al.(2000)Auer, Cesa{-}Bianchi, Freund, and Schapire]{Auer1998}
Peter Auer, Nicol{\`{o}} Cesa{-}Bianchi, Yoav Freund, and Robert~E. Schapire.
\newblock Gambling in a rigged casino: The adversarial multi-armed bandit problem.
\newblock \emph{Electron. Colloquium Comput. Complex.}, {TR00-068}, 2000.

\bibitem[Auer et~al.(2002)Auer, Cesa{-}Bianchi, Freund, and Schapire]{Auer03}
Peter Auer, Nicol{\`{o}} Cesa{-}Bianchi, Yoav Freund, and Robert~E. Schapire.
\newblock The nonstochastic multiarmed bandit problem.
\newblock \emph{{SIAM} J. Comput.}, 32\penalty0 (1):\penalty0 48--77, 2002.

\bibitem[Bart{\'{o}}k et~al.(2014)Bart{\'{o}}k, Foster, P{\'{a}}l, Rakhlin, and Szepesv{\'{a}}ri]{Bartok14}
G{\'{a}}bor Bart{\'{o}}k, Dean~P. Foster, D{\'{a}}vid P{\'{a}}l, Alexander Rakhlin, and Csaba Szepesv{\'{a}}ri.
\newblock Partial monitoring - classification, regret bounds, and algorithms.
\newblock \emph{Math. Oper. Res.}, 39\penalty0 (4):\penalty0 967--997, 2014.

\bibitem[Cesa{-}Bianchi et~al.(2006)Cesa{-}Bianchi, Lugosi, and Stoltz]{Cesa06}
Nicol{\`{o}} Cesa{-}Bianchi, G{\'{a}}bor Lugosi, and Gilles Stoltz.
\newblock Regret minimization under partial monitoring.
\newblock \emph{Math. Oper. Res.}, 31\penalty0 (3):\penalty0 562--580, 2006.

\bibitem[Chaudhuri and Tewari(2017)]{Chaudhuri17}
Sougata Chaudhuri and Ambuj Tewari.
\newblock Online learning to rank with top-k feedback.
\newblock \emph{J. Mach. Learn. Res.}, 18:\penalty0 103:1--103:50, 2017.

\bibitem[Kleinberg and Leighton(2003)]{Kleinberg03}
Robert~D. Kleinberg and Frank~Thomson Leighton.
\newblock The value of knowing a demand curve: Bounds on regret for online posted-price auctions.
\newblock In \emph{44th Symposium on Foundations of Computer Science 2003)}, pages 594--605. {IEEE} Computer Society, 2003.

\bibitem[Lattimore(2022)]{Lattimore22}
Tor Lattimore.
\newblock Minimax regret for partial monitoring: Infinite outcomes and rustichini's regret.
\newblock In \emph{Conference on Learning Theory}, volume 178 of \emph{Proceedings of Machine Learning Research}, pages 1547--1575. {PMLR}, 2022.

\bibitem[Lattimore and Gy{\"{o}}rgy(2021)]{Lattimore021a}
Tor Lattimore and Andr{\'{a}}s Gy{\"{o}}rgy.
\newblock Mirror descent and the information ratio.
\newblock In \emph{Conference on Learning Theory, {COLT} 2021}, volume 134 of \emph{Proceedings of Machine Learning Research}, pages 2965--2992. {PMLR}, 2021.

\bibitem[Lattimore and Szepesv{\'{a}}ri(2018)]{Lattimore18arXiv}
Tor Lattimore and Csaba Szepesv{\'{a}}ri.
\newblock Cleaning up the neighborhood: {A} full classification for adversarial partial monitoring.
\newblock \emph{CoRR}, abs/1805.09247, 2018.

\bibitem[Lattimore and Szepesv{\'{a}}ri(2019{\natexlab{a}})]{Lattimore18}
Tor Lattimore and Csaba Szepesv{\'{a}}ri.
\newblock Cleaning up the neighborhood: {A} full classification for adversarial partial monitoring.
\newblock In \emph{Algorithmic Learning Theory, {ALT} 2019}, volume~98 of \emph{Proceedings of Machine Learning Research}, pages 529--556. {PMLR}, 2019{\natexlab{a}}.

\bibitem[Lattimore and Szepesv{\'{a}}ri(2019{\natexlab{b}})]{LattimoreS19}
Tor Lattimore and Csaba Szepesv{\'{a}}ri.
\newblock An information-theoretic approach to minimax regret in partial monitoring.
\newblock In \emph{Conference on Learning Theory, {COLT} 2019}, volume~99 of \emph{Proceedings of Machine Learning Research}, pages 2111--2139. {PMLR}, 2019{\natexlab{b}}.

\bibitem[Lattimore and Szepesv{\'{a}}ri(2020)]{LattimoreS20}
Tor Lattimore and Csaba Szepesv{\'{a}}ri.
\newblock Exploration by optimisation in partial monitoring.
\newblock In \emph{Conference on Learning Theory, {COLT} 2020}, volume 125 of \emph{Proceedings of Machine Learning Research}, pages 2488--2515. {PMLR}, 2020.

\bibitem[Liu(2011)]{Liu11}
Tie{-}Yan Liu.
\newblock \emph{Learning to Rank for Information Retrieval}.
\newblock Springer, 2011.

\bibitem[Piccolboni and Schindelhauer(2001)]{Piccolboni2001}
Antonio Piccolboni and Christian Schindelhauer.
\newblock Discrete prediction games with arbitrary feedback and loss.
\newblock In \emph{Computational Learning Theory, 14th Annual Conference on Computational Learning Theory, {COLT} 2001 and 5th European Conference on Computational Learning Theory, EuroCOLT 2001}, volume 2111 of \emph{Lecture Notes in Computer Science}, pages 208--223. Springer, 2001.

\bibitem[Suehiro et~al.(2012)Suehiro, Hatano, Kijima, Takimoto, and Nagano]{Suehiro12}
Daiki Suehiro, Kohei Hatano, Shuji Kijima, Eiji Takimoto, and Kiyohito Nagano.
\newblock Online prediction under submodular constraints.
\newblock In \emph{Algorithmic Learning Theory - 23rd International Conference, {ALT} 2012}, volume 7568 of \emph{Lecture Notes in Computer Science}, pages 260--274. Springer, 2012.

\bibitem[Tsuchiya et~al.(2020)Tsuchiya, Honda, and Sugiyama]{TsuchiyaHS20}
Taira Tsuchiya, Junya Honda, and Masashi Sugiyama.
\newblock Analysis and design of thompson sampling for stochastic partial monitoring.
\newblock In \emph{Advances in Neural Information Processing Systems 33: Annual Conference on Neural Information Processing Systems 2020, NeurIPS 2020}, 2020.

\end{thebibliography}

\newpage



\appendix

\section{Proofs for Section \ref{sec:finite_partial_monitoring}}

\textbf{Proof of Lemma \ref{lem:alter_go_lo}.}
\begin{proof}
We first prove that the alternative definition of global observability (Definition \ref{defn:alter_go_lo}) generalizes the original definition (Definition \ref{defn:go}). It will follow that the alternative definition of local observability (Definition \ref{defn:alter_go_lo}) generalizes the original definition (Definition \ref{defn:lo}).

Consider top-$k$ feedback model, so each signal matrix $S_{k'}$ is $2^{k}$ by $2^m$. Definition \ref{defn:go} says that a pair of actions $\sigma_i$ and $\sigma_j$ is globally observable if $l_i - l_j \in \oplus_{1 \le k' \le m!} \col(S_{k'}^\top)$. So we can write $l_i - l_j$ as
\[
l_i - l_j = \sum_{k'=1}^{m!} \sum_{l'=1}^{2^k} c_{k', l'} (S_{k'}^\top)_{l'} \, ,
\]
where $c_{k', l'} \in \R$ is constant, and $(S_{k'}^\top)_{l'}$ is the $l'$-th column of $S_{k'}^\top$. Define
\begin{align}
f(\sigma_{k'}, H_{k', k''}) = \sum_{l'=1}^{2^k} c_{k', l'} (S_{k'}^\top)_{k'', l'} \quad \text{for } 1 \le k'' \le 2^m \, ,
\label{eq:a1}
\end{align}
where $(S_{k'}^\top)_{k'', l'}$ is the element in row $k''$ and column $l'$ of $S_{k'}^\top$. Then $l_i - l_j$ can also be written as
\[
l_i - l_j = \sum_{k'=1}^{m!}
\begin{bmatrix}
f(\sigma_{k'}, H_{k', 1}) \\
... \\
f(\sigma_{k'}, H_{k', 2^m})
\end{bmatrix} \, ,
\]
satisfying Definition \ref{defn:alter_go_lo}.

Now, we prove the other direction. That is, the original definition of global observability (Definition \ref{defn:go}) generalizes the alternative definition (Definition \ref{defn:alter_go_lo}). It will follow that the original definition of local observability (Definition \ref{defn:lo}) generalizes the alternative definition (Definition \ref{defn:alter_go_lo}).

Definition \ref{defn:alter_go_lo} says that a pair of actions $\sigma_i$ and $\sigma_j$ is globally observable if there exists a function $f : \Sigma \times \mathcal{H} \to \R$ such that 
\begin{align}
    \sum_{k'=1}^{m!} f(\sigma_{k'}, H_{k',k''}) = L_{i,k''} - L_{j, k''} \quad \text{for all } 1 \le k'' \le 2^m \, . \label{eq:a2}
\end{align}
Note that in this setting (online learning with top-$k$ feedback), $\mid \Sigma \mid = m!$ and $\mid \mathcal{H} \mid = 2^k$, so function $f$ has a finite domain. Using the definition of signal matrices (Definition \ref{defn:signal-matrix}), we see that in Eq. \eqref{eq:a1},
\[
(S_{k'}^\top)_{k'', l'} = (S_{k'})_{l', k''} = \I(H_{k',k''} = s_{l'}) \, ,
\]
where $s_{l'}$ is the $l'$-th symbol in $\mathcal{H}$. So, among all $(S_{k'}^\top)_{k'', l'}$, $1 \le l' \le 2^m$, exactly one of them is $1$. Therefore, we can let $c_{k', l'} = f(\sigma_{k'}, H_{k',k''})$ for $l'$ such that $H_{k',k''} = s_{l'}$, and $c_{k', l'} = 0$ otherwise.
Then, Eq. \eqref{eq:a2} can be written as
\[
    L_{i,k''} - L_{j, k''} =  \sum_{k'=1}^{m!} \sum_{l'=1}^{2^k} c_{k', l'} (S_{k'}^\top)_{k'', l'} \quad \text{for all } 1 \le k'' \le 2^m \, .
\]
So, 
\[
l_i - l_j = \sum_{k'=1}^{m!}
\begin{bmatrix}
\sum_{l'=1}^{2^k} c_{k', l'} (S_{k'}^\top)_{1, l'} \\
... \\
\sum_{l'=1}^{2^k} c_{k', l'} (S_{k'}^\top)_{2^m, l'}
\end{bmatrix}
= \sum_{k'=1}^{m!} \sum_{l'=1}^{2^k} c_{k', l'} (S_{k'}^\top)_{l'} \, ,
\]
satisfying Definition \ref{defn:go}.
\end{proof}

\section{Proofs for Section \ref{sec:pl_sl_dcg}}

\textbf{Proof of Lemma \ref{lem:rl_pareto_optimal_actions}.}
\begin{proof}
For $p \in \Delta$, $l_i \cdot p = \sum_{j=1}^{2^m} p_j (f(\sigma_{i}) \cdot R_j) = f(\sigma_{i}) \cdot \sum_{j=1}^{2^m} p_j R_j = f(\sigma_{i}) \cdot \E[R]$, where the expectation is taken with respect to $p$. Since $f(\sigma_i)$ is an element-wise strictly increasing transformation of $\sigma_{i}^{-1}$, then $l_i \cdot p$ is minimized when $\E[R(\sigma_i(1))] \ge \E[R(\sigma_i(2))] \ge ... \ge \E[R(\sigma_i(m))]$. Therefore, the cell of $\sigma_i$ is $C_i = \{ p \in \Delta : \ones^\top p = 1, \E[R(\sigma_i(1))] \ge \E[R(\sigma_i(2))] \ge ... \ge \E[R(\sigma_i(m))] \}$.
Note that $C_i$ is non-empty.
$C_i$ has only one equality constraint and hence has dimension $(2^m - 1)$. This shows $\sigma_i$ is Pareto-optimal.
\end{proof}

\textbf{Proof of Lemma \ref{lem:rl_neighboring_action_pairs}.}
\begin{proof}
We first prove the ``if'' part.

From Lemma \ref{lem:rl_pareto_optimal_actions}, we know each of learner's actions $\sigma_i$ is Pareto-optimal and it has cell
\begin{multline*}
C_i = \{ p \in \Delta : \E[R(\sigma_i(1))]  \ge ... \ge \E[R(\sigma_i(k))] \\
\ge ... \ge \E[R(\sigma_i(k'))] \ge ... \ge \E[R(\sigma_i(m))] \} \, .
\end{multline*}
Let $\sigma_i(k) = a$, $\sigma_i(k+1)=b$, $\sigma_j(k+1)=a$ and $\sigma_j(k)=b$ for some $k \in [m-1]$. For all $k' \in [m] \setminus \{k,k+1\}$, let $\sigma_i(k') = \sigma_j(k')$. Then
\begin{align*}
C_i \cap C_j &= \{ p \in \Delta : \E[R(\sigma_i(1))]  \ge ... \ge \E[R(\sigma_i(k))] \\
&= \E[R(\sigma_i(k+1))] \ge ... \ge \E[R(\sigma_i(m))] \} \, .
\end{align*}
Therefore, there are two equalities in $C_i \cap C_j$: $\ones^\top p = 1$ and $\sum_{j=1}^{2^m} p_j (R_j(\sigma_i(k)) - R_j(\sigma_i(k+1))) = 0$. These two equations define two non-parallel hyperplanes, so the intersection of the two hyperplanes has dimension $(2^m - 2)$. By Definition \ref{defn:neighbors}, $\sigma_i$ and $\sigma_j$ are neighboring actions.

We then prove the ``only if'' part.

Assume the condition (\textit{there is exactly ...}) fails to hold. Note that $\sigma_i$ and $\sigma_j$ cannot be the same, so they differ in at least two positions. Then there are two cases. 1. $\sigma_i$ and $\sigma_j$ differ in exactly two positions, but the two positions are not consecutive. 2. $\sigma_i$ and $\sigma_j$ differ in more than two positions.

Consider case 1. Without loss of generality, assume $\sigma_i$ and $\sigma_j$ differ in positions $k$ and $k'$ where $k' - k > 1$. Then there is a unique pair of objects $\{a,b\}$ such that $\sigma_i(k) = a$, $\sigma_i(k') = b$, $\sigma_j(k) = b$ and $\sigma_j(k') = a$, and $\sigma_i(l) = \sigma_j(l)$ for $l \neq k, k'$. From Lemma \ref{lem:rl_pareto_optimal_actions}, $\sigma_i$ has cell
\begin{align*}
C_i = \{ p \in \Delta : & \E[R(\sigma_i(1))]  \ge ... \ge \E[R(\sigma_i(k))] \ge ... \\
& \ge \E[R(\sigma_i(k'))] \ge ... \ge \E[R(\sigma_i(m))] \} \, ,
\end{align*}
and $\sigma_j$ has cell
\begin{align*}
C_j = \{ p \in \Delta : & \E[R(\sigma_j(1))]  \ge ... \ge \E[R(\sigma_j(k))] \ge ... \\
& \ge \E[R(\sigma_j(k'))] \ge ... \ge \E[R(\sigma_j(m))] \} \, .
\end{align*}
Then
\begin{align*}
C_i \cap C_j = \{ p \in \Delta : 
&\E[R(\sigma_i(1))]  \ge ... \ge \E[R(\sigma_i(k))] \ge ...\\
&\ge \E[R(\sigma_i(k'))] \ge ... \ge \E[R(\sigma_i(m))],\\
&\E[R(\sigma_j(1))]  \ge ... \ge \E[R(\sigma_j(k))] \ge ...\\
&\ge \E[R(\sigma_j(k'))] \ge ... \ge \E[R(\sigma_j(m))] \} \, .
\end{align*}
Since
\begin{align*}
\E[R(\sigma_i(k))] = \E[R(\sigma_j(k'))] = \E[R(a)] \, ,
\end{align*}
and
\begin{align*}
\E[R(\sigma_i(k'))] = \E[R(\sigma_j(k))] = \E[R(b)] \, ,
\end{align*}
it follows that $C_i \cap C_j$ has a constraint
\begin{align*}
\E[R(\sigma_i(k))] = ... = \E[R(\sigma_i(k'))] \, .
\end{align*}
Since $k' - k > 1$, $\E[R(\sigma_i(k))] = ... = \E[R(\sigma_i(k'))]$ has at least two equalities. Then $C_i \cap C_j$ has at least three equality constraints (including $\ones^\top p = 1$), which shows $C_i \cap C_j$ has dimension less than $(2^m - 2)$. Therefore, $\{ \sigma_i, \sigma_j \}$ is not a neighboring action pair.

Now consider case 2. If $\sigma_i$ and $\sigma_j$ differ in more than two positions, then there are at least two pairs of objects such that for each pair, the relative order of the two objects in $\sigma_i$ is different from that in $\sigma_j$. Applying the argument for case 1 to case 2 shows $\{ \sigma_i, \sigma_j \}$ is not a neighboring action pair.
\end{proof}

\textbf{Proof of Lemma \ref{lem:rl_neighborhood_action_set}.}
\begin{proof}
By definition of neighborhood action set, $N_{i,j}^{+} = \{k : 1 \le k \le m!, C_i \cap C_j \subseteq C_k \}$. \citet{Bartok14} mentions that if $N_{i,j}^{+}$ contains some other action $\sigma_k$, then either $C_k = C_i$, $C_k = C_j$, or $C_k = C_i \cap C_j$. From Lemma \ref{lem:rl_pareto_optimal_actions}, for $RL$ each of learner's actions is Pareto-optimal, so $\dim(C_k) = 2^m - 1$. This shows $C_k \neq C_i \cap C_j$. To see $C_k \neq C_i$, assume for contraction that $C_k = C_i$. Then this means that both actions $\sigma_i$ and $\sigma_k$ are optimal under $p$, $\forall p \in C_k$, which implies $0 = l_i \cdot p - l_k \cdot p = p \cdot (l_i - l_k)$ for all $p \in C_k$. For $RL$, $l_i \neq l_k$ for $i \neq k$. Moreover, $l_i - l_k$ is independent of $\ones$ for $i \neq k$ because each action is Pareto-optimal. Then $p \cdot (l_i - l_k) = 0$ for all $p \in C_k$ would impose another equality constraint on $C_k$, so $\dim(C_k) \leq 2^m - 2$. We know $\dim(C_k) = 2^m - 1$, a contraction. This shows $C_k \neq C_i$. Similarly, we have $C_k \neq C_j$. Therefore, $N_{i,j}^{+} = \{i,j\}$ and $\oplus_{k \in N_{i,j}^{+}} \col(S_{k}^\top) = \col(S_{i}^\top) \oplus \col(S_{j}^\top)$.
\end{proof}

\textbf{Proof of Theorem \ref{thm:rl}.}
\begin{proof}
\textbf{Part 1:} We first prove the local observability fails for $k = 1,...,m-2$. It suffices to show the local observability fails for $k = m-2$ because top-$k$ feedback has strictly more information than top $k'$ feedback does for $k' < k$.

Note that for the signal matrix for the top-$k$ feedback model (defined according to Definition \ref{defn:signal-matrix}) when $k = m-2$, each row has exactly 4 ones and each column has exactly 1 one.

Consider two actions $\sigma_1 = 1,2,3,...,m-2,m-1,m$ and $\sigma_2 = 1,2,3,...,m-2,m,m-1$. That is, $\sigma_1$ gives object $i$ rank $i$ for $1 \le i \le m$. $\sigma_2$ gives object $i$ rank $i$ for $1 \le i \le m-2$, object $m$ rank $m-1$ and object $m-1$ rank $m$. By Lemma \ref{lem:rl_neighboring_action_pairs}, $\sigma_1$ and $\sigma_2$ are neighboring actions.

Inspired by observations from Remark \ref{rem:rl_loss_diff_observation}, we form $2^{m-2}$ groups of $4$ relevance vectors such that within each group, the relevance vectors only differ at object $m-1$ and $m$. Correspondingly, we divide the vector $l_1 - l_2$ into $2^{m-2}$ groups. Then each group is $[0,\ f^s(m) - f^s(m-1),\ f^s(m-1) - f^s(m),\ 0]$ (see Remark \ref{rem:rl_loss_diff_observation}). For signal matrices $S_1$ and $S_2$, we can also form $2^{m-2}$ groups of $4$ columns accordingly. For $k=m-2$, the signal matrix is of size $2^{m-2} \times 2^m$, and in this case, $\sigma_1$ and $\sigma_2$ have the same signal matrix $S = S_1 = S_2$ because $\sigma_1$ and $\sigma_2$ have exactly the same feedback no matter what the relevance vector is. Now in each group, there are only two types of rows of $S$, namely $[0\ 0\ 0\ 0]$ and $[1\ 1\ 1\ 1]$. Table \ref{rl:table1} shows $l_1 - l_2$ and two types of rows of $S$ for each group. Since $f^s$ is strictly increasing, it is clear that $l_1 - l_2 \notin \col(S^\top)$. This shows the local observability fails for $k=m-2$.
\begin{table}[ht]
\centering
\begin{tabular}{|c|c|c|c|c|}
\hline
 & $R(m-1) = 0$ & $R(m-1) = 0$ & $R(m-1) = 1$ & $R(m-1) = 1$\\
 & $R(m) = 0$ & $R(m) = 1$ & $R(m) = 0$ & $R(m) = 1$\\\hline
$l_1-l_2$ & 0 & $f^s(m) - f^s(m-1)$ & $f^s(m-1) - f^s(m)$ & 0\\\hline
\multirow{2}{*}{rows of $S$} & 1 & 1 & 1 & 1\\
& 0 & 0 & 0 & 0\\\hline
\end{tabular}
\caption{Part 1, within the group, $R(c)$ is the same for all $c \neq m-1, m$. See proof of Theorem \ref{thm:rl} for more details.}
\label{rl:table1}
\end{table}

\textbf{Part 2:} We then prove the local observability holds for $k = m-1, m$. Again, it suffices to show the local observability holds for $k = m-1$. (Note that for $k = m$, the game has bandit feedback, and thus is locally observable as in Section 2.1 of \citet{Bartok14}.)

Note that for the signal matrix when $k = m-1$, each row has exactly 2 ones and each column has exactly 1 one.

Consider neighboring action pair $\{\sigma_i, \sigma_j\}$. Let $\{a,b\}$ be a pair of objects as in Remark \ref{rem:rl_loss_diff_observation}. We proceed similarly as in Part 1. We form $2^{m-2}$ groups of $4$ relevance vectors such that within each group, the relevance vectors only differ at object $a$ and $b$. Correspondingly, we divide the vector $l_i - l_j$ into $2^{m-2}$ groups. Then each group is $[0,\ f^s(k'+1) - f^s(k'),\ f^s(k') - f^s(k'+1),\ 0]$. For signal matrices $S_i$ and $S_j$, we can also form $2^{m-2}$ groups of $4$ columns accordingly. Then there are two cases:\\
(1) Neither $a$ nor $b$ is ranked last by $\sigma_i$ or $\sigma_j$, so the relevance for $a$ and the relevance for $b$ are both revealed through feedback. Concatenate $S_i$ and $S_j$ by row and denote the resultant matrix by $S$. $S$ is of size $2^{m} \times 2^m$. Now in each group, there are only five types of rows of $S$, as shown in Table \ref{rl:table2}. It is clear that the piece $[0,\ f^s(k'+1) - f^s(k'),\ f^s(k') - f^s(k'+1),\ 0]$ is in the row space of $S$. In this case, $l_i - l_j \in \col(S_{i}^\top) \oplus \col(S_{j}^\top)$.\\
(2) Either $a$ or $b$ is ranked last by $\sigma_i$ or $\sigma_j$, so only one of the relevance for $a$ and the relevance for $b$ is revealed through feedback. Concatenate $S_i$ and $S_j$ by row and denote the resultant matrix by $S$. $S$ is of size $2^{m} \times 2^m$. Now in each group, there are only five types of rows of $S$, as shown in Table \ref{rl:table3}. The piece $[0,\ f^s(k'+1) - f^s(k'),\ f^s(k') - f^s(k'+1),\ 0]$ is in the row space of $S$ because $[0,\ f^s(k'+1) - f^s(k'),\ f^s(k') - f^s(k'+1),\ 0] = (f^s(k'+1) - f^s(k'))[1\ 1\ 0\ 0] - (f^s(k'+1) - f^s(k'))[1\ 0\ 1\ 0]$. In this case, $l_i - l_j \in \col(S_{i}^\top) \oplus \col(S_{j}^\top)$.\\
\begin{table}[ht]
\centering
\begin{tabular}{|c|c|c|c|c|}
\hline
 & $R(a) = 0$ & $R(a) = 0$ & $R(a) = 1$ & $R(a) = 1$\\
 & $R(b) = 0$ & $R(b) = 1$ & $R(b) = 0$ & $R(b) = 1$\\\hline
$l_i-l_j$ & 0 & $f^s(k'+1) - f^s(k')$ & $f^s(k') - f^s(k'+1)$ & 0\\\hline
\multirow{5}{*}{rows of $S$} & 1 & 0 & 0 & 0\\
& 0 & 1 & 0 & 0\\
& 0 & 0 & 1 & 0\\
& 0 & 0 & 0 & 1\\
& 0 & 0 & 0 & 0\\\hline
\end{tabular}
\caption{Part 2 (1), within the group, $R(c)$ is the same for all $c \neq a, b$. See proof of Theorem \ref{thm:rl} for more details.}
\label{rl:table2}
\end{table}
\begin{table}[ht]
\centering
\begin{tabular}{|c|c|c|c|c|}
\hline
 & $R(a) = 0$ & $R(a) = 0$ & $R(a) = 1$ & $R(a) = 1$\\
 & $R(b) = 0$ & $R(b) = 1$ & $R(b) = 0$ & $R(b) = 1$\\\hline
$l_i-l_j$ & 0 & $f^s(k'+1) - f^s(k')$ & $f^s(k') - f^s(k'+1)$ & 0\\\hline
\multirow{5}{*}{rows of $S$} & 1 & 1 & 0 & 0\\
& 0 & 0 & 1 & 1\\
& 1 & 0 & 1 & 0\\
& 0 & 1 & 0 & 1\\
& 0 & 0 & 0 & 0\\\hline
\end{tabular}
\caption{Part 2 (2), within the group, $R(c)$ is the same for all $c \neq a, b$. See proof of Theorem \ref{thm:rl} for more details.}
\label{rl:table3}
\end{table}

In either case, we have $l_i - l_j \in \col(S_{i}^\top) \oplus \col(S_{j}^\top)$, so $\{\sigma_i, \sigma_j\}$ is locally observable. Hence the local observability holds, concluding the proof.
\end{proof}

\section{Proofs for Section \ref{sec:p@n}}

\textbf{Proof of Lemma \ref{lem:p@n_pareto_optimal_actions}.}
\begin{proof}
The negated P@n is defined as $-P@n(\sigma, R) = f(\sigma) \cdot R$ where $f(\sigma) = [-\I(\sigma^{-1}(1) \le n), ..., -\I(\sigma^{-1}(m) \le n)]$. For any $p \in \Delta$, we have $l_i \cdot p = \sum_{j=1}^{2^m} p_j (f(\sigma_{i}) \cdot R_j) = f(\sigma_{i}) \cdot (\sum_{j=1}^{2^m} p_j R_j) = f(\sigma_{i}) \cdot \E[R]$, where the expectation is taken with respect to $p$. Let $A_i = \{ a : \I(\sigma_{i}^{-1}(a) \le n)  = 1 \}$ and $B_i = \{ b : \I(\sigma_{i}^{-1}(b) \le n)  = 0 \}$ be subsets of $\{1,2,...,m\}$. $A_i$ is the set of objects contributing to the loss while $B_i$ is the set of objects not contributing to the loss. Then $l_i \cdot p$ is minimized when the expected relevances of objects are such that $\E[R(a)] \ge \E[R(b)]$ for all $a \in A_i, b \in B_i$. Therefore, $C_i = \{p \in \Delta: \ones^\top p = 1, \E[R(a)] \ge \E[R(b)], \forall a \in A_i, \forall b \in B_i\}$. $C_i$ has only one equality constraint and hence has dimension $(2^m - 1)$. This shows action $\sigma_i$ is Pareto-optimal.
\end{proof}

\textbf{Proof of Lemma \ref{lem:p@n_neighboring_action_pairs}.}
\begin{proof}
For the ``if'' part, assume the condition (\textit{there is exactly ...}) holds. From Lemma \ref{lem:p@n_pareto_optimal_actions}, action $\sigma_i$ is Pareto-optimal and its cell is $C_i = \{p \in \Delta: \E[R(x)] \ge \E[R(y)], \forall x \in A_i, y \in B_i \}$. Action $\sigma_j$ is also Pareto-optimal and its cell is $C_j = \{p \in \Delta: \E[R(x)] \ge \E[R(y)], \forall x \in A_j, y \in B_j \}$. Then $C_i \cap C_j = \{p \in \Delta: \E[R(a)] = \E[R(b)] \text{ and } \E[R(x)] \ge \E[R(y)], \forall x \in A_i, y \in B_i \text{ and } \E[R(z)] \ge \E[R(w)], \forall z \in A_j, w \in B_j \}$. $C_i \cap C_j$ has only two equality constraints (counting $\ones^\top p = 1$), and hence it has dimension $(2^m - 2)$. Therefore, $\{\sigma_i, \sigma_j\}$ is a neighboring action pair.

For the ``only if'' part, assume the condition (\textit{there is exactly ...}) does not hold. Note that for negated P@n, $\mid A_i\mid  = n$ and $\mid B_i\mid  = m-n$ for all action $\sigma_i$. There are two cases. 1. $\mid A_i \setminus A_j\mid  = 0$. 2. $\mid A_i \setminus A_j\mid  > 1$.

For the first case, if $\mid A_i \setminus A_j\mid  = 0$, then $A_i = A_j$ and $B_i = B_j$. Then $C_i \cap C_j = C_i$ has dimension $(2^m - 1)$ because $\sigma_i$ is Pareto-optimal by Lemma \ref{lem:p@n_pareto_optimal_actions}. Thus, in this case, $\{\sigma_i, \sigma_j\}$ is not a neighboring action pair.

For the second case, if $\mid A_i \setminus A_j\mid  > 1$, then there are at least two pair of objects $\{a,b\}$ and $\{a',b'\}$ such that $a,a' \in A_i$, $a,a' \in B_j$, $b,b' \in B_i$, and $b,b' \in A_j$. Following the arguments in the ``if'' part, it is easy to show that $C_i \cap C_j$ has at least three equality constraints (counting $\ones^\top p = 1$), and hence it has dimension less than $(2^m - 2)$. Thus, in this case, $\{\sigma_i, \sigma_j\}$ is not a neighboring action pair.
\end{proof}

\textbf{Proof of Lemma \ref{lem:p@n_neighborhood_action_set}.}
\begin{proof}
By definition of neighborhood action set, $N_{i,j}^{+} = \{k : 1 \le k \le m!, C_i \cap C_j \subseteq C_k \}$. \citet{Bartok14} mentions that if $N_{i,j}^{+}$ contains some other action $\sigma_k$, then either $C_k = C_i$, $C_k = C_j$, or $C_k = C_i \cap C_j$. From Lemma \ref{lem:p@n_pareto_optimal_actions}, every action is Pareto-optimal for negated P@n, so $\dim(C_k) = 2^m - 1$. Hence $C_k \neq C_i \cap C_j$. If $C_k = C_i$, then both actions $\sigma_i$ and $\sigma_k$ are optimal under $p$, $\forall p \in C_k$, which implies $0 = l_i \cdot p - l_k \cdot p = p \cdot (l_i - l_k)$ for all $p \in C_k$. Since $C_k$ has dimension $(2^m-1)$, $p \cdot (l_i - l_k) = 0$ cannot impose an equality constraint on $C_k$. Therefore, $l_i=l_k$. Similarly, if $C_k = C_j$, then $l_j = l_k$. This shows $N_{i,j}^{+} = \{k: 1 \le k \le m!, l_k = l_i \text{ or } l_k = l_j\}$.
\end{proof}

\textbf{Proof of Theorem \ref{thm:p@n}.}
\begin{proof}
It suffices to show the local observability holds for $k=1$ because there is strictly more information for the game with $k > 1$ than that with $k=1$.

Note that for the signal matrix when $k = 1$, each row has exactly $2^{m-1}$ ones and each column has exactly 1 one.

Consider neighboring action pair $\{\sigma_i, \sigma_j\}$. Let $\{a,b\}$ be a pair of objects as in Remark \ref{rem:p@n_loss_diff_observation}. We form $2^{m-2}$ groups of $4$ relevance vectors such that within each group, the relevance vectors only differ at object $a$ and $b$. Correspondingly, we divide the vector $l_i - l_j$ into $2^{m-2}$ groups. Then each group is $[0\ 1\ -1\ 0]$. For signal matrices $S_l$ where $l \in N_{i,j}^{+}$, we can also form $2^{m-2}$ groups of $4$ columns accordingly. Then concatenate all $2n!(m-n)!$ \footnote{See Remark \ref{rem:p@n_duplicate_actions} for how this number is calculated.} signal matrices $S_l$ where $l \in N_{i,j}^{+}$ by row and denote the resultant matrix by $S$. $S$ is of size $4n!(m-n)! \times 2^m$. Now in each group, there are only five types of rows of $S$, as shown in Table \ref{p@n:table1}. $[1\ 1\ 0\ 0]$ and $[0\ 0\ 1\ 1]$ correspond to the action $\sigma_l$ with $l \in N_{i,j}^{+}$ that puts object $a$ rank $1$. $[1\ 0\ 1\ 0]$ and $[0\ 1\ 0\ 1]$ correspond to the action $\sigma_{l'}$ with $l' \in N_{i,j}^{+}$ that puts object $b$ rank $1$. The piece $[0\ 1\ -1\ 0]$ is in the row space of $S$ because $[0\ 1\ -1\ 0] = 2[1\ 1\ 0\ 0] + [0\ 0\ 1\ 1] - 2[1\ 0\ 1\ 0] - [0\ 1\ 0\ 1]$. Therefore, $l_i - l_j \in \oplus_{l \in N_{i,j}^{+}} \col(S_{l}^\top)$, so $\{\sigma_i, \sigma_j\}$ is locally observable and the local observability holds for P@n.
\begin{table}
\centering
\begin{tabular}{|c|c|c|c|c|}
\hline
 & $R(a) = 0$ & $R(a) = 0$ & $R(a) = 1$ & $R(a) = 1$\\
 & $R(b) = 0$ & $R(b) = 1$ & $R(b) = 0$ & $R(b) = 1$\\\hline
$l_i-l_j$ & 0 & 1 & -1 & 0\\\hline
\multirow{5}{*}{rows of $S$} & 1 & 1 & 0 & 0\\
& 0 & 0 & 1 & 1\\
& 1 & 0 & 1 & 0\\
& 0 & 1 & 0 & 1\\
& 0 & 0 & 0 & 0\\\hline
\end{tabular}
\caption{P@n, within the group, $R(c)$ is the same for all $c \neq a, b$. See proof of Theorem \ref{thm:p@n} for more details.}
\label{p@n:table1}
\end{table}
\end{proof}

\newpage

\section{Proofs for Section \ref{sec:p@n_algo}}

In this section, the loss function is negated P@n unless otherwise stated. We consider top-1 feedback model as described in Section \ref{sec:p@n_algo}.
\\

\textbf{Proof of Lemma \ref{lem:nw2_V}.}
\begin{proof}
Lemma \ref{lem:p@n_neighborhood_action_set} shows for neighboring action pair $\{a, b\}$, the neighborhood action set is $N_{a,b}^{+} = \{k: 1 \le k \le m!, l_k = l_a \text{ or } l_k = l_b\}$ where $l_a$ and $l_b$ are loss vectors of actions $a$ and $b$ respectively.

For top-1 feedback model, each signal matrix $S_{k'}$ is $2$ by $2^m$. By definition of local observability (Definition \ref{defn:lo}), we can write $l_a - l_b$ as
\[
l_a - l_b = \sum_{k' \in N_{a,b}^{+}} \Big[ c_{k', 1} (S_{k'}^\top)_{1} + c_{k', 2} (S_{k'}^\top)_{2} \Big] \, ,
\]
where $c_{k', l'} \in \R$ is constant, and $(S_{k'}^\top)_{l'}$ is the $l'$-th column of $S_{k'}^\top$, for $l' = 1,2$. Define
\begin{align}
v^{ab}(\sigma_{k'}, H_{k', k''}) = \Big[ c_{k', 1} (S_{k'}^\top)_{k'', 1} + c_{k', 2} (S_{k'}^\top)_{k'', 2} \Big], \quad \text{for } 1 \le k'' \le 2^m \, ,
\label{eq:def_v_ab}
\end{align}
where $(S_{k'}^\top)_{k'', l'}$ is the element in row $k''$ and column $l'$ of $S_{k'}^\top$, for $l'=1,2$. Then $l_a - l_b$ can also be written as
\[
l_a - l_b =\sum_{k' \in N_{a,b}^{+}}
\begin{bmatrix}
v^{ab}(\sigma_{k'}, H_{k', 1}) \\
... \\
v^{ab}(\sigma_{k'}, H_{k', 2^m})
\end{bmatrix} \, .
\]
Now back to Equation (\ref{eq:def_v_ab}), $(S_{k'}^\top)_{k'', 1}$ and $(S_{k'}^\top)_{k'', 2}$ are binary for all $k', k''$. From the proof for Theorem \ref{thm:p@n}, we can choose $c_{k',1}$ and $c_{k',2}$ such that $\mid c_{k',1}\mid  \le 2$ and $\mid c_{k',2}\mid  \le 2$ for all $k'$. Then it follows that $\lVert v^{ab} \rVert_\infty = \max_{\sigma \in \Sigma, s \in \mathcal{H}} \mid v^{ab} (\sigma, s)\mid  \le 4$, completing the proof.
\end{proof}

\textbf{Proof of Lemma \ref{lem:nw2_eps_G}.}
\begin{proof}
We follow the proof for Lemma 5 in \citet{Lattimore18} with some modifications to ensure $\frac{1}{\epsilon_G}$ is in the order of $\mathrm{poly}(m)$.

Since $u \in C_d$, we have $(l_c - l_d) \cdot u \ge 0$. The result is trivial if $c, d$ are neighbors or $(l_c - l_d) \cdot u = 0$.\\
\begin{figure}[h]
\centering
\vspace{-0.2cm}
\begin{tikzpicture}[scale=0.8]
\draw[thin] (0,0) -- (3,0) -- (0,3) -- (0,0);
\draw[thin] (0,1.5) -- (1.5,1.5);
\draw[thin] (1.5,0) -- (1.5,1.5);
\draw[fill=black] (2.3,0.3) circle (1pt);
\draw[fill=black] (0.5,2) circle (1pt);
\node (Ca) at (0.3,2.3) {\footnotesize {$C_c$}};
\node (Cta) at (1.8,0.3) {\footnotesize {$C_d$}};
\node (Cb) at (0.3,0.3) {\footnotesize {$C_e$}};
\node (v) at (1.3,2.5) {$v$};
\node (u) at (3.1,0.8) {$u$};
\node (w) at (1.9,2.2) {$w$};
\draw[->,thin,shorten >=2pt] (v) -- (0.5,2);
\draw[->,thin,shorten >=2pt] (u) -- (2.3,0.3);
\draw[->,thin,shorten >=2pt] (w) -- (1.02,1.5);
\draw[densely dotted,thin,<-,shorten <=2pt,shorten >=2pt] (2.3,0.3) -- (0.5,2);
\draw[fill=black] (1.02,1.5) circle (1pt);
\end{tikzpicture}
\caption{Illustrating proof for Lemma \ref{lem:local_regret}, adopted from \citet{Lattimore18}.}
\label{fig:1}
\end{figure}

Now assume $c, d$ are not neighbors and $(l_c - l_d) \cdot u > 0$. Let $v$ be the centroid of $C_c$. Consider the line segment connecting $u$ and $v$. Then let $w$ be the first point on this line segment for which there exists $e \in N_c \cap \mathcal{A}$ with $w \in C_e$ (see Figure \ref{fig:1}). $w$ is well-defined by the Jordan-Brouwer separation theorem, and $e$ is well-defined because $\mathcal{A}$ is a duplicate-free set of Pareto-optimal classes.

Recall that each class $c' \in \mathcal{A}$ corresponds to a unique partition of $[m]$ into two subsets $A_{c'}$ and $B_{c'}$ such that only objects in $A_{c'}$ contribute to the calculation of negated P@n. For each $c' \in \mathcal{A}$, we can define $f(c') = [\I(1 \in A_{c'}), ..., \I(m \in A_{c'})]$. Let $\mathbf{R} = [R_1, ..., R_{2^m}]$ collect all relevance vectors: the $i$-th column of $\mathbf{R}$ is the $i$-th relevance vector $R_i$. Then we can rewrite $(l_{c'} - l_{d'}) \cdot u'$ as
\[
(l_{c'} - l_{d'}) \cdot u' = (-f(c') \cdot \mathbf{R} + f(d') \cdot \mathbf{R}) \cdot u' = (-f(c') + f(d')) \cdot \mathbf{R} u' \, ,
\]
for all $c', d' \in \mathcal{A}$ and $u' \in \Delta$. Note that $\mathbf{R} u' = \E_{u'} [R]$ is the expected relevance vector under $u'$.

Now, using twice $(l_c - l_e) \cdot w = 0$, we calculate
\begin{align}
\begin{split}
(l_c - l_e) \cdot u
&= (l_c - l_e) \cdot (u-w) \\
&= (-f(c) + f(e)) \cdot \mathbf{R} (u-w) \\
&= \frac{\norm{\mathbf{R} (u-w)}_2}{\norm{\mathbf{R} (w-v)}_2} (-f(c) + f(e)) \cdot \mathbf{R} (w-v) \\
&= \frac{\norm{\mathbf{R} (u-w)}_2}{\norm{\mathbf{R} (w-v)}_2} (l_c - l_e) \cdot (w-v) \\
&= \frac{\norm{\mathbf{R} (u-w)}_2}{\norm{\mathbf{R} (w-v)}_2} (l_e - l_c) \cdot v > 0
\end{split}
\label{eq:1}
\end{align}
where the third equality uses that $w \ne v$ is a point of the line segment connecting $v$ and $u$, so that $w-v$ and $u-w$ are parallel and have the same direction. Note that $(l_e - l_c) \cdot v > 0$ because $c,e$ are different Pareto-optimal classes and $v$ is the centroid of $C_c$. $\norm{\mathbf{R} (w-v)}_2 = \norm{\E_w [R] - \E_v [R]}_2 > 0$ because otherwise, $\E_w [R] = \E_v [R]$ would imply $(l_c - l_e) \cdot v = (-f(c)+f(e)) \cdot \E_v[R] = (-f(c)+f(e)) \cdot \E_w[R] = 0$, contradicting that $v$ is the centroid of $C_c$. To see $\norm{\mathbf{R} (u-w)}_2 > 0$, we recalculate $(l_c - l_e) \cdot u$ in another way
\begin{align}
(l_c - l_e) \cdot u &= (l_c - l_e) \cdot (u-w) \nonumber \\
&= \frac{\norm{u-w}_2}{\norm{w-v}_2} (l_c - l_e) \cdot (w-v) \nonumber \\
&= \frac{\norm{u-w}_2}{\norm{w-v}_2} (l_e - l_c) \cdot v > 0 \, .
\label{eq:2}
\end{align}
The inequality in Equation (\ref{eq:2}) holds because $\norm{u-w}_2 > 0$ (since $(l_c - l_d) \cdot u > 0$) and $\norm{w-v}_2 > 0$. Therefore, $\norm{\mathbf{R} (u-w)}_2 > 0$ in Equation (\ref{eq:1}) also holds.

Let $v_{c'}$ be the centroid of $C_{c'}$ for any $c' \in \mathcal{A}$. Then we have
\begin{align*}
\frac{(l_c - l_d) \cdot u}{(l_c - l_e) \cdot u}
&
=\frac{(l_c - l_d) \cdot (w+u-w)}{(l_c - l_e) \cdot u} \\
&
\stackrel{\text{\footnotesize (a)}}{<}
\frac{(l_c - l_e) \cdot w + (l_c - l_d) \cdot (u-w)}{(l_c - l_e) \cdot u} \\
&
\stackrel{\text{\footnotesize (b)}}{=}
\frac{(l_c - l_d) \cdot (u-w)}{(l_c - l_e) \cdot u} \\
&
= \frac{(-f(c)+f(d)) \cdot \mathbf{R} (u-w)}{(-f(c) + f(e)) \cdot \mathbf{R} u} \\
&
\stackrel{\text{\footnotesize (c)}}{=}
\frac{\norm{\mathbf{R} (w-v)}_2 (-f(c)+f(d)) \cdot \mathbf{R} (u-w)}{\norm{\mathbf{R} (u-w)}_2 (l_e - l_c) \cdot v} \\
&
\stackrel{\text{\footnotesize (d)}}{\le}
\frac{\norm{\mathbf{R} (w-v)}_2 \norm{-f(c)+f(d)}_2}{(l_e - l_c) \cdot v} \\
&
= \frac{\norm{\E_w [R] - \E_v [R]}_2 \norm{-f(c)+f(d)}_2}{(l_e - l_c) \cdot v} \\
&
\stackrel{\text{\footnotesize (e)}}{\le}
\frac{m}{\min_{c' \in \mathcal{A}} \min_{d' \in N_{c'} \cap \mathcal{A}} (l_{d'} - l_{c'}) \cdot v_{c'}}
\end{align*}
where (a) follows since $(l_c - l_d) \cdot w < 0 = (l_c - l_e) \cdot w$, (b) follows since $(l_c - l_e) \cdot w = 0$, (c) follows by Equation (\ref{eq:1}), (d) follows by Cauchy-Schwarz. Note that $0 \preceq \E [R] \preceq 1$, we can bound $\norm{\E_w [R] - \E_v [R]}_2$ by $\sqrt{m}$. Since both $f(c)$ and $f(d)$ are binary vectors, we can bound $\norm{-f(c)+f(d)}_2$ by $\sqrt{m}$. Then (e) follows since $v$ is the centroid of $C_c$ and $(l_e - l_c) \cdot v \ge \min_{c' \in \mathcal{A}} \min_{d' \in N_{c'} \cap \mathcal{A}} (l_{d'} - l_{c'}) \cdot v_{c'}$ ($v_{c'}$ is the centroid of $C_{c'}$).

Finally, we want to find a lower bound for
\[
\min_{c' \in \mathcal{A}} \min_{d' \in N_{c'} \cap \mathcal{A}} (l_{d'} - l_{c'}) \cdot v_{c'} = \min_{c' \in \mathcal{A}} \min_{d' \in N_{c'} \cap \mathcal{A}} (-f(d')+f(c')) \cdot \E_{v_{c'}} [R] \, .
\]
Note that for any $c' \in \mathcal{A}$, $\E_{v_{c'}} [R(i)] = 1$ if object $i \in A_{c'}$ and $\frac{1}{2}$ otherwise (by the symmetry property of the centroid $v_{c'}$ of $C_{c'}$). Along with observations from Remark \ref{rem:p@n_loss_diff_observation}, we have
\[
(-f(d')+f(c')) \cdot \E_{v_{c'}} [R] = \frac{1}{2} \, ,
\]
for all $c' \in \mathcal{A}$ and $d' \in N_{c'} \cap \mathcal{A}$. Therefore, we can bound
\begin{align}
\frac{m}{\min_{c' \in \mathcal{A}} \min_{d' \in N_{c'} \cap \mathcal{A}} (l_{d'} - l_{c'}) \cdot v_{c'}} \le 2m := \frac{1}{\epsilon_G} \, .
\label{eq:def_epsilon_G}
\end{align}
$\frac{1}{\epsilon_G}$ is clearly a polynomial of $m$.
\end{proof}


\end{document}